\documentclass[letterpaper]{article} 
\usepackage{aaai25}  
\usepackage{times}  
\usepackage{helvet}  
\usepackage{courier}  
\usepackage[hyphens]{url}  
\usepackage{enumerate}
\usepackage{enumitem}
\usepackage{graphicx, calc} 
\usepackage{natbib}  
\usepackage{caption} 
\usepackage{algorithm}
\usepackage{algorithmic}

\usepackage{booktabs}

\usepackage[accsupp]{axessibility}  %
\usepackage[dvipsnames]{xcolor}
\usepackage{mathtools}
\usepackage{multirow}
\usepackage{blindtext}
\usepackage{amsmath}
\usepackage{amssymb}
\usepackage[capitalize]{cleveref}
\usepackage{tabularray}
\usepackage{colortbl}
\usepackage{subfigure}
\usepackage{subcaption}
\usepackage{amssymb}
\usepackage{pifont}
\newcommand{\cmark}{\ding{51}}%
\newcommand{\xmark}{\ding{55}}%
\usepackage{float}

\newlength\myheight
\newlength\mydepth
\settototalheight\myheight{Xygp}
\usepackage{color, colortbl}
\definecolor{Gray}{gray}{0.9}

\usepackage{newfloat}
\usepackage{listings}
\DeclareCaptionStyle{ruled}{labelfont=normalfont,labelsep=colon,strut=off} 
\floatstyle{ruled}
\newfloat{listing}{tb}{lst}{}
\floatname{listing}{Listing}

\title{OmniCount: Multi-label Object Counting with Semantic-Geometric Priors}
\author{Anindya Mondal$^1$$^2$$^3$\thanks{Authors have equal contributions.} , Sauradip Nag$^1$$^2$$^5$\footnotemark[1], Xiatian Zhu$^1$$^2$$^3$, Anjan Dutta$^1$$^2$$^3$$^4$ \\
$^1$University of Surrey, $^2$CVSSP, $^3$Surrey Institute for People-Centred AI, \\ $^4$School of Veterinary Medicine, $^5$iFlyTek-Surrey Joint Research Center on AI\\
{\tt\small \{a.mondal, s.nag, xiatian.zhu, anjan.dutta\}@surrey.ac.uk,}}

\newcommand{\eg}{\emph e.g. }
\newcommand{\ie}{\emph i.e. }
\newcommand{\myparagraph}[1]{\vspace{0pt}\noindent{\bf #1}}

\definecolor{skyblue}{rgb}{0.53, 0.81, 0.92}
\definecolor{skybluehighlight}{rgb}{0.53, 0.81, 0.92}

\begin{document}
\maketitle

\begin{abstract}

Object counting is pivotal for understanding the composition of scenes. Previously, this task was dominated by class-specific methods, which have gradually evolved into more adaptable class-agnostic strategies. However, these strategies come with their own set of limitations, such as the need for manual exemplar input and multiple passes for multiple categories, resulting in significant inefficiencies. This paper introduces a more practical approach enabling simultaneous counting of multiple object categories using an open-vocabulary framework. Our solution, OmniCount, stands out by using semantic and geometric insights (priors) from pre-trained models to count multiple categories of objects as specified by users, all without additional training. OmniCount distinguishes itself by generating precise object masks and leveraging varied interactive prompts via the Segment Anything Model for efficient counting. To evaluate OmniCount, we created the OmniCount-191 benchmark, a first-of-its-kind dataset with multi-label object counts, including points, bounding boxes, and VQA annotations. Our comprehensive evaluation in OmniCount-191, alongside other leading benchmarks, demonstrates OmniCount's exceptional performance, significantly outpacing existing solutions. The project webpage is available at https://mondalanindya.github.io/OmniCount.
\end{abstract}

\section{Introduction}
\label{sec:intro}

Understanding object distribution across multiple categories is crucial for comprehensive scene analysis, driving increased interest in object counting research. It aims to estimate specific object counts in natural scenes.
Traditionally, object counting has focused on class-specific methods for categories such as human crowds \cite{li2018csrnet,song2021rethinking,haniccv2023steerer,Li2023CalibCrowd,Liang2023CrowdCLIP,liu2023pet}, cells \cite{Khan2016EmbryonicCell}, fruits \cite{Rahnemoonfar2017DeepCount}, and vehicles \cite{Bui2020VehicleCount}. However, these methods require extensive training data and are limited to predefined categories. 
Recent efforts have shifted towards class-agnostic counting, using exemplars (cropped images and class names) to count arbitrary categories \cite{Chattopadhyay2017CountEveryday, Ranjan2021CountEverything, Ranjan2022ExemplarFree, Jiang2023CLIPCount}. Some operate in low-shot settings \cite{you2022fewshot,xu2023zero}, but they still require substantial training data and separate processing for each category, increasing computational demands for multi-category scenes.
Further, detection and instance segmentation methods \cite{Chattopadhyay2017CountEveryday,Cholakkal2022PartialSupervision} can count multiple categories by name, but struggle with small or non-atomic objects like grapes or bananas, which are hard to detect individually.

\begin{figure*}[!t]
    \centering
    \includegraphics[width=\linewidth]{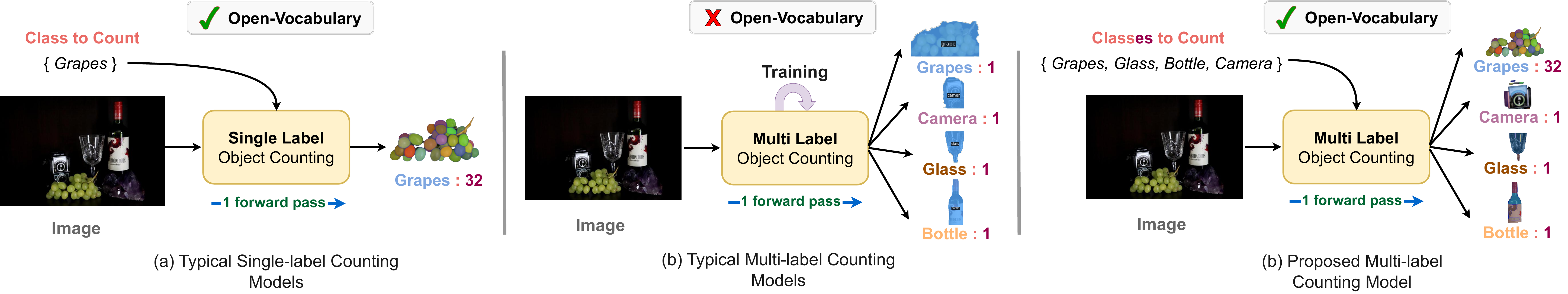}
    \caption{\textbf{Object counting paradigms:} (a) Typical single-label object counting models support open-vocabulary counting but processes only {\em a single category} one time. (b) Existing multi-label object counting models are training-based (i.e., not open-vocabulary) approaches and also fail to count {\em non-atomic objects} (\eg, grapes). (c) We advocate more efficient and convenient {\em multi-label open-vocabulary counting} that is training-free, and supports counting all the target categories in a single pass.} 
    \label{fig:enter-label}
\vspace{-15pt}
\end{figure*}
Inspired by these observations, we introduce \textbf{OmniCount}, a novel method for efficiently counting multiple open-vocabulary object categories simultaneously in a single forward pass. Unlike prior zero-shot object counting \cite{xu2023zero,Dai2024REC} requiring substantial training on labelled seen categories, 
this proposed open-vocabulary object counting method leverages vision-language models to count objects across a broad spectrum of categories without any training (\ie, training-free).
OmniCount distinguishes itself by utilizing semantic and geometric cues from pre-trained foundation models to partition images into semantically coherent regions, identify occluded objects with depth cues, and ensure precise object delineation. A key feature is its use of geometric cues for object recovery and reducing overcounting.
In general, segmentation models typically struggle with dense scenarios, leading to hallucination effects where distant or occluded objects get missed \cite{xie2204m2bev,philion2020lift}. 
Extracting semantic and geometric priors separately helps us achieve the best of both domains. Specifically, our model uses metric depth for rectifying dense scenes where semantic estimation fails. By performing $k$-nearest neighbors-based searches, we refine or even recover overlooked instances, utilizing recovered object features to estimate reference points per class, enabling the counting model to handle similar-looking objects (see \cref{fig:reference_point}). This enables the detection of objects of varying shapes, sizes, and densities, facilitating leveraging segmentation models like the Segment Anything Model (SAM) \cite{kirillov2023segment} to generate individual object masks. SAM's ability to use points as segmentation prompts for fine-grained, non-atomic object segmentation makes it our preferred module for counting.


As an under-studied area, object counting lacks a dataset with a diverse range of annotations for multiple generic categories per image. The sparsely populated object detection datasets like PASCAL VOC \cite{hoiem2009pascal} and MS COCO \cite{lin2014microsoft} cannot adequately represent real-world counting challenges. Additionally, the recently proposed REC-8K dataset \cite{Dai2024REC} focuses on fine-grained counting, distinguishing objects within the same category, like ``red apples'' vs. ``green apples'' but it doesn't support counting across different coarse-grained categories. 
The absence of a suitable dataset for this underexplored domain prompted the creation of the \textbf{OmniCount-191} benchmark. This comprehensive dataset includes 302,300 object instances across 191 categories in 30,230 images, featuring multiple categories per image and a variety of detailed annotations such as counts, points and bounding boxes for each object (\cref{fig:dataset_diss}).

We make the following contributions:
(1) We re-promote multi-label object counting that bypasses the conventional reliance on object detection and semantic segmentation models, addressing common accuracy issues such as over- and under-counting;
(2) We introduce a novel, efficient, and user-friendly framework \textbf{OmniCount} for multi-label object counting by leveraging semantic and geometric cues without necessitating additional training; 
(3) We create a new multi-label object counting dataset, \textbf{OmniCount-191}, with rich annotations for fostering the development of this newly introduced setting; 
(4) We conduct extensive experiments to demonstrate OmniCount's superior performance over existing methods on our dataset and establish benchmarks.

\section{Related Work}
\label{sec:related}
\myparagraph{Learning-based object counting:} Traditional counting methods have focused on specific categories like crowds \cite{li2018csrnet, song2021rethinking, haniccv2023steerer, huang2023counting, Liang2023CrowdCLIP, liu2023pet, Peng2024SingleDomainCrowdCount, Guo2024MutualCrowdCount}, cells \cite{Khan2016EmbryonicCell}, fruits \cite{Rahnemoonfar2017DeepCount}, and vehicles \cite{Bui2020VehicleCount}, mainly using regression-based techniques to create density maps from point annotations \cite{lempitsky2010learning, zhang2016single, xu2021crowd}. These methods rely on point annotations to generate density maps, which train models that predict object counts by summing pixel values in the predicted density map. This class-specific approach is effective for its trained categories but lacks the flexibility for broader applications involving multiple object categories. 
In contrast, class-agnostic counting aims for versatility, using exemplars to count objects of any category \cite{lu2019class,zhang2019nonlinear,Ranjan2021CountEverything,shi2022represent,gupta2021visual,zhang2021look, Ranjan2022ExemplarFree,shi2023focus}. Some data-efficient variants operate in zero-shot \cite{xu2023zero,xupami2023zero,Jiang2023CLIPCount, Dai2024REC} and few-shot \cite{you2022fewshot,yang2021class} settings, trained on seen or base classes to handle unseen or novel categories. 
These methods use similarity maps for flexible counting across classes, but learning-based models require extensive data, making them difficult to scale. We propose an open-vocabulary object counter that counts using prompts like points, boxes, or text, eliminating the need for training and expanding possibilities for diverse scenarios without the data and training burden.

\myparagraph{Multi-label object counting:} Despite the advancements in single-label counting, real-world scenarios often involve scenes with multiple object classes coexisting \cite{you2022fewshot}. Prior works by \cite{cholakkal2019object,Cholakkal2022PartialSupervision} and \cite{Chattopadhyay2017CountEveryday} have explored multi-label counting in sparse settings, focusing on global counts and labels within human discernible ranges. However, these methods struggle to identify non-atomic or densely clustered objects, such as grapes. Few-shot counting \cite{Ranjan2021CountEverything} attempts to address these but typically restricts to one category per image. Recently, \cite{Dai2024REC} introduced GrREC, a model for counting multiple fine-grained categories, but it requires training in predefined seen categories. They also developed the REC-8K dataset with images and corresponding referring expressions. In contrast, our open-vocabulary model uses semantic and geometric cues from pre-trained models without additional training. Moreover, we emphasize the need for datasets capturing real-world use cases and dense, multi-class interactions, leading to the creation of OmniCount-191.

\begin{figure*}[!t]
    \centering
    \includegraphics[width=0.9\linewidth]{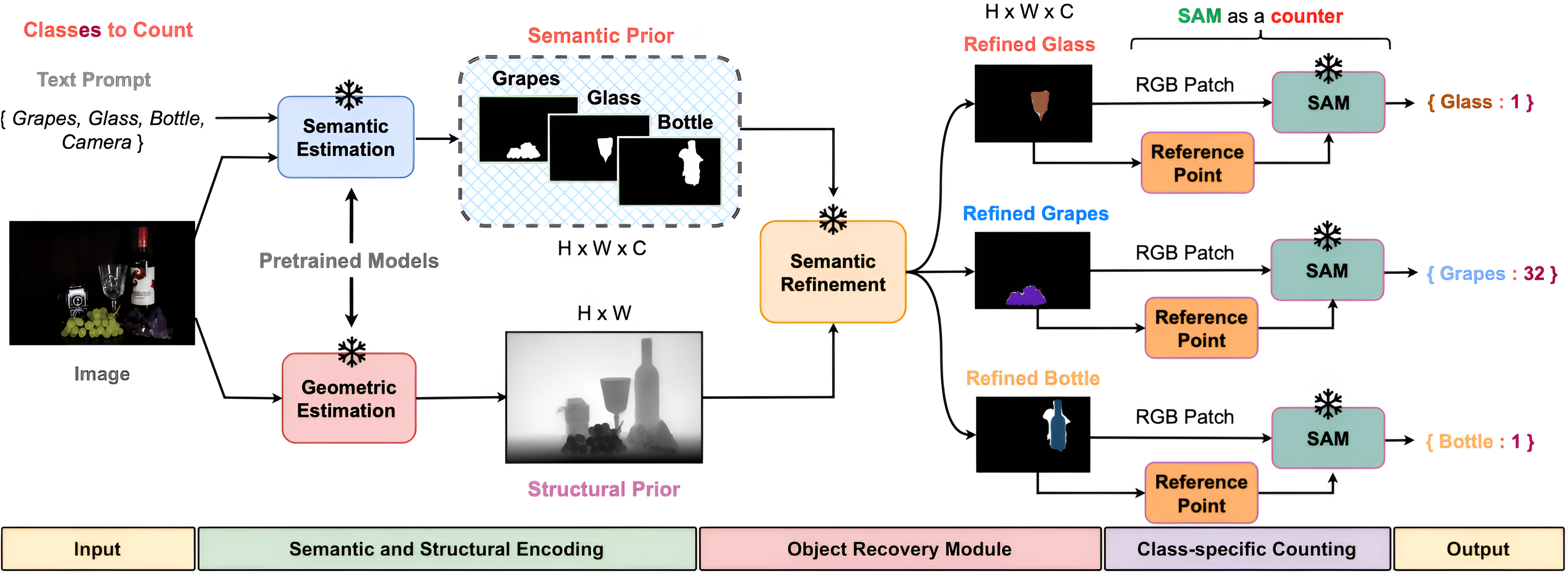}
    \caption{\textbf{OmniCount pipeline:} OmniCount processes the input image and target object classes using Semantic Estimation (SAN) and Geometric Estimation (Marigold) modules to generate class-specific masks and depth maps. These initial semantic and geometric priors are then refined through an Object Recovery module, 
    producing precise binary masks. The refined masks help extract RGB patches and reference points, reducing over-counting. SAM then uses these RGB patches and reference points to generate instance-level masks, resulting in accurate object counts. (\raisebox{-\mydepth}{\fbox{\includegraphics[height=\myheight]{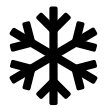}}} denotes pre-trained, frozen models)}
    \label{fig:pipeline}
    \vspace{-14pt}
\end{figure*}

\myparagraph{Prompt-based foundation models:} 
LLMs like GPT \cite{brown2020language} have transformed NLP and computer vision, excelling in zero-shot and few-shot tasks. Foundation models like CLIP \cite{radford2021learning} use contrastive learning to align text and image, enabling effective low-shot transfer through textual prompts.
In image segmentation, the Segment Anything Model (SAM) \cite{kirillov2023segment} generates precise object masks from diverse prompts (points, boxes, text), excelling in various benchmarks and showcasing robust zero-shot abilities. An ideal object counter should be \textit{visually promptable}, \textit{interactive}, and capable of \textit{open-set} counting. While SAM possesses these traits, it struggles with occlusions \cite{ji2023sam} and multi-class object counting \cite{shi2024training} due to its class-agnostic approach. We address these challenges by incorporating depth and semantic priors into SAM, enhancing its effectiveness for complex counting tasks involving occlusions and multiple object classes, thus approaching the ideal counting model.

\vspace{-0.2in}
\section{OmniCount}
\label{sec:model}


In this work, we aim to achieve open-vocabulary, multi-label training-free object counting within a given image and with a set of labels to be counted in that image. Our proposed model is illustrated in \cref{fig:pipeline}. 
\vspace{-0.1in}
\subsection{Problem formulation}
\label{sec:prob}
The problem of multi-label object counting can be defined as obtaining an object counter $\mathcal{F}_\text{count}$ using a training set $\mathcal{D}_\text{train}=\{(I_1, \mathcal{P}_1, \mathcal{C}_1), \ldots, (I_N, \mathcal{P}_N, \mathcal{C}_N)\}$, where each $I_i \in \mathbb{R}^{H \times W \times 3}$ represents an RGB image, $\mathcal{P}_i=\{p_1, \ldots, p_{m_i}\}$ is a set of class labels and $\mathcal{C}_i=\{c_1, \ldots, c_{m_i}\}$ are the corresponding object counts (\ie object with label $p_k$ occurs $c_k$ times in $I_i$), with $m_i$ being the number of unique objects in the $i$-th image and $N$ the total number of training data points in $\mathcal{D}_\text{train}$. 
For an image $I_k$ and a subset of labels $\{p_1,\ldots,p_{k_l}\} \subseteq \mathcal{P}_k$, the function $\mathcal{F}_\text{count}$ should result in: 
%
\begin{equation}
    \{c_1,\ldots,c_{k_l}\} = \mathcal{F}_\text{count}(I_k, \{p_1,\ldots,p_{k_l}\})
\end{equation}
where $c_{k_l}$ is the number of occurrences of the object with label $p_{k_l}$ in the image $I_k$.
Our goal is to develop an open-vocabulary multi-label object counting model $\mathcal{F}_\text{count}$, such that it generalizes well to $\mathcal{D}_\text{test}$, a held-out test set of data points with classes not in $\mathcal{D}_\text{train}$, \ie, $\mathcal{D}_\text{train} \cap \mathcal{D}_\text{test} = \phi$.
To achieve this, we introduce OmniCount, a multi-label object counting model that utilizes semantic and geometric priors, avoiding training that requires large datasets and expensive computational resources. 
Since our model is training-free, we do not use $\mathcal{D}_\text{train}$ and only evaluate our model on $\mathcal{D}_\text{test}$.

\subsection{Semantic and structural encoding}

\label{encoding}
\myparagraph{Semantic estimation module:} 
To count multiple objects in a single forward pass, we segment the image into relevant semantic regions. While any standard open-vocabulary segmentation model 
can be used, we employ the Side Adapter Network (SAN) \cite{xu2023side} as a semantic segmentation model $\mathcal{E}_\text{sem}$ that takes an image $I$ and a set of class labels $\mathcal{P}=\{p_1,\ldots,p_m\}$ as input and results in $\textbf{S}_\mathcal{P}=\{ S_1, \ldots, S_m\}$, a set of binary semantic masks corresponding to the classes in $\mathcal{P}$ as follows:
\begin{equation}
\label{eq:2}
\mathbf{S}_\mathcal{P}, F_\mathcal{P} = \mathcal{E}_\text{sem}(I, \mathcal{P})
\end{equation}
where $F_\mathcal{P} \in \mathbb{R}^{\frac{H}{K}\times\frac{W}{K}\times C}$ is an intermediate low-resolution feature activations, with $K$ and $C$ being integers depending on the design of $\mathcal{E}_\text{sem}$. We use $\mathbf{S}_\mathcal{P}$ as a semantic prior, bridging the gap between multi-label counting and semantic awareness.
While segmenting 2D RGB images, SAM \cite{kirillov2023segment} primarily relies on texture information, such as colour. Combined with occlusion, this reliance can result in over-segmented masks (see \cref{fig:pipeline} for the ``bottle’’ class) and, consequently, over-counting. To address this, we incorporate geometric information to achieve fine-grained segmentation, mitigating over-segmentation and over-counting. We refine the segmentation mask using geometric priors to be discussed in the next paragraph. 

\myparagraph{Geometric estimation module:} Classical segmentation models \cite{Long2015FCN,kirillov2023segment} primarily rely on texture information for object delineation, which often fails under significant occlusion. 
So, counting dense interacting or overlapping objects requires information beyond RGB.
Therefore, similar to how density maps have been utilized in classical object counting by providing a clearer representation of object distribution, we leverage depth maps to enhance segmentation accuracy.
Depth maps, like density maps, ignore texture and focus on structural information, aiding in the segmentation of objects regardless of their distance from the camera, as shown in \cref{fig:pipeline}. This structural prior helps recover hidden objects and refine object semantics.
Using an off-the-shelf depth map rendering model Marigold \cite{ke2023repurposing} denoted by $\mathcal{E}_\text{depth}$, we generate depth map $D$ for the image $I$ as \(D = \mathcal{E}_\text{depth}(I)\), which serve as geometric prior for our model.
Notably, $D$ is a matrix of the same spatial dimension as the image $I$, with pixels having normalized depth values in $[0,1]$. This implies that for a pixel $x \in I$, $D(x)\in[0,1]$ indicates the normalized depth value of $x$.
We utilize this pixel-wise depth information to refine the coarse segmentation mask $\mathbf{S}_\mathcal{P}$, discussed in the following subsection.

\begin{figure}[!t]
    \centering
    \includegraphics[width=0.8\linewidth]{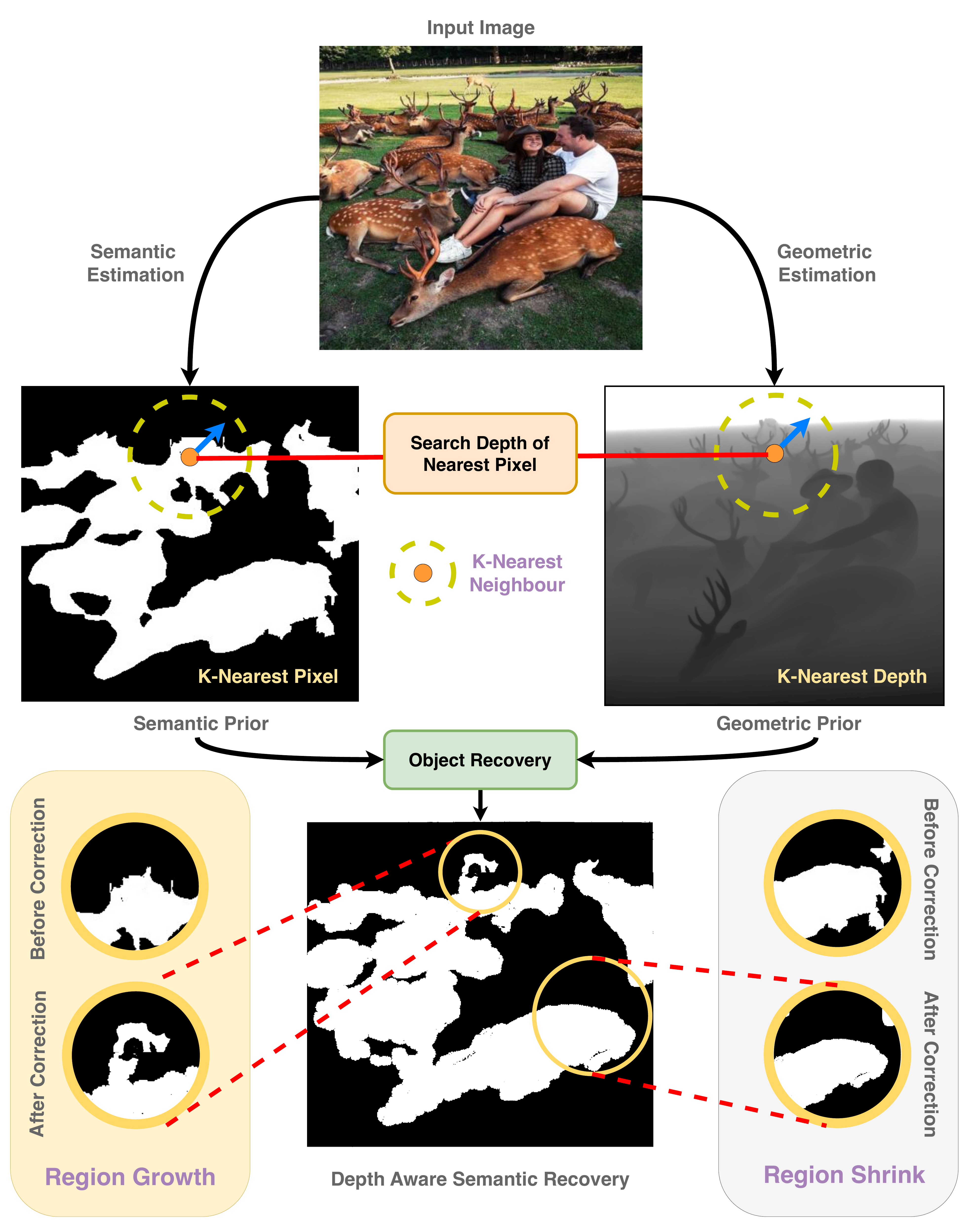}
    \caption{\textbf{Geometry aware Object Recovery:} We refine semantic masks with geometric priors using k-nearest neighbor searches to filter edge pixels by category uniqueness and depth alignment, enhancing mask precision through depth-integrated segmentation.}
    \label{fig:refinement_depth}
\end{figure}

\subsection{Geometry aware object recovery}
\label{refinement}
\begin{figure*}[!hbtp]
\centering
    \includegraphics[width=0.9\textwidth]{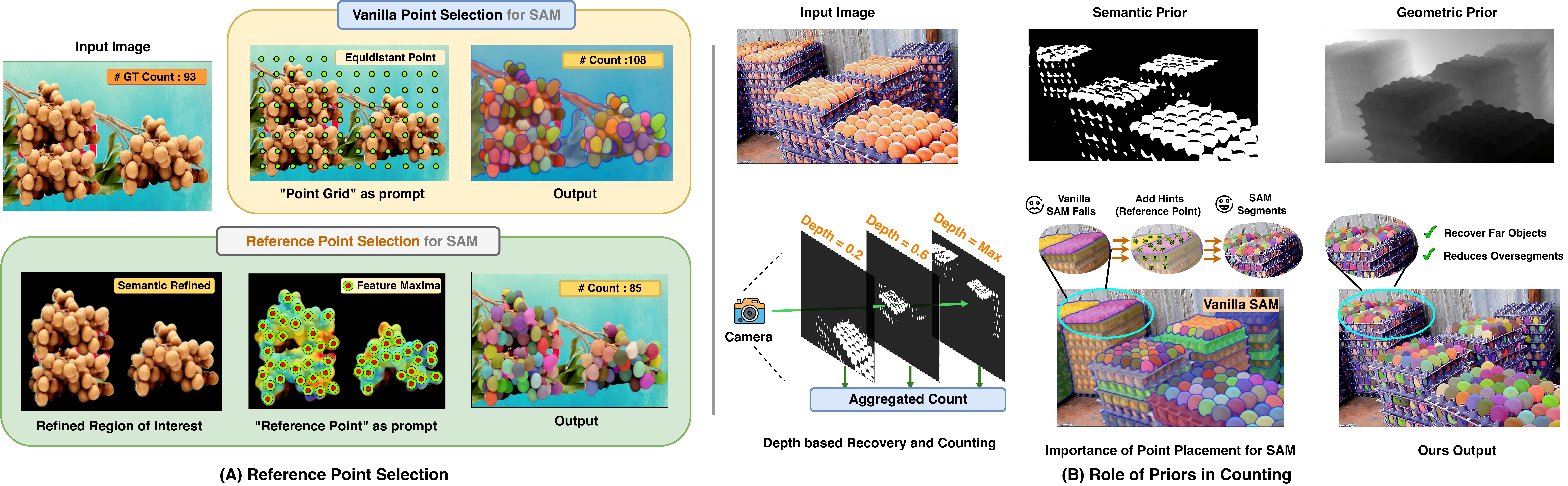}
    \caption{\textbf{Reference Point Selection:} SAM’s segmentation accuracy is enhanced by refining reference point selection. Panel (A) shows how integrating semantic priors, identifying local maxima, and applying Gaussian refinement improve reference point accuracy, focusing them on foreground objects for better segmentation and counting. Panel (B) demonstrates the benefits of incorporating semantic and geometric priors, where depth-based recovery and precise reference points help SAM recover distant or occluded objects, reducing over-segmentation issues found in the default ``\textit{everything mode}''.}
    \label{fig:reference_point}
    \vspace{-14pt}
\end{figure*}

\begin{figure}[!ht]
\centering
    \includegraphics[width=0.8\columnwidth, keepaspectratio]{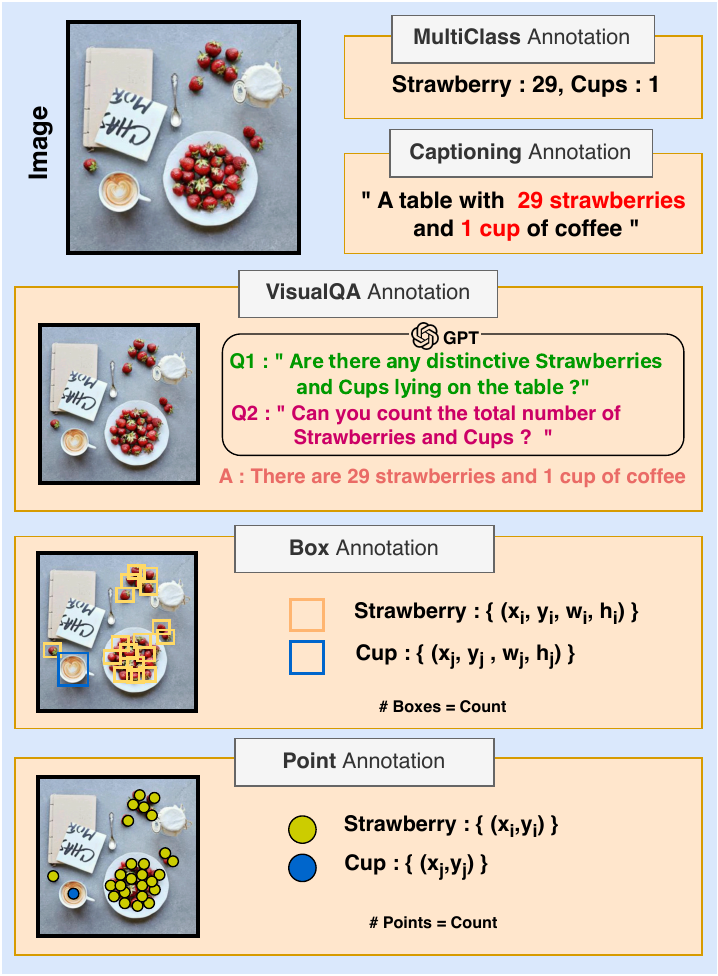}
    \caption{\textbf{OmniCount-191 Annotations:} A collection of images with $191$ classes across nine domains, annotating each image with captions, VQA, boxes, and points.}
    \label{fig:dataset_diss}
\end{figure}

Providing precise reference points is crucial for guiding SAM towards accurate counting and minimizing over-counting. We inject geometric priors into semantic priors before passing them into SAM (see \cref{fig:reference_point}) to obtain such reference points. 
To refine the semantic prior $\mathbf{S}_\mathcal{P}$ with geometric insights from the depth prior $D$, we conduct a k-nearest neighbor (kNN) (refer to \cref{fig:refinement_depth}) search centering around each of the pixels of the individual semantic mask to ensure two conditions are met:\\
\noindent (1) A pixel must exclusively belong to its designated object category, preventing any overlapping with masks of other categories. For a pixel $x$ in mask $S_j$:
\begin{equation*}
C_1(x): x \notin S_k, \forall k \neq j
\end{equation*}
(2) The absolute difference between a pixel's depth and the mean depth of its object category must fall below a specified tolerance $\tau$, ensuring depth consistency (for objects with curved edges). For a pixel $x$ in $S_n$ with mean depth $D_\mu(S_n)$:
\begin{equation*}
C_2(x): | D(x) - D_\mu(S_n) | < \tau
\end{equation*}
where $D(x)$ denotes the depth of pixel $x$ and $D_\mu(S_n)=\frac{1}{s}\sum_y D(y), \forall y\in S_n$ with $s$ as the total number of pixels in $S_n$. Pixels that fulfil both conditions $C_1(x)$ and $C_2(x)$ are integrated into their appropriate object class, leading to a refined semantic prior $\mathbf{S}'_\mathcal{P}$, computed as \(\mathbf{S}'_\mathcal{P} = \mathcal{R}_\text{geom}(\mathbf{S}_\mathcal{P}, D)\), where $\mathcal{R}_\text{geom}$ is the geometry-aware semantic refinement function, enhancing the precision of semantic masks by considering depth information. This depth-aware refined mask (see \cref{fig:refinement_depth}) minimizes the risk of over-segmentation (see \cref{table:ablations}) or recovers undiscovered objects in occluded scenes.

\subsection{Reference point guided counting}
\label{sam module}
We use SAM \cite{kirillov2023segment} as an object counter, which employs a point grid generator to place uniform points across the image and generate masks. This can lead to overcounting due to points falling on both foreground and background. To prevent this, we propose a reference point selection procedure that focuses solely on the foreground.

\myparagraph{Reference point selection:} 
We select reference points (refer to \cref{fig:reference_point}) for the SAM decoder using the feature activation $F_\mathcal{P}$ from semantic priors (see \cref{eq:2}) to enhance text-image similarity accuracy.
A set of reference points $\mathbf{P}=\{ \rho_1, \ldots, \rho_s \}$ are identified as local maxima within $F_\mathcal{P}$, but direct upsampling can misalign \cite{zhang2020distribution} them due to quantization errors (refer to \cref{table:ablations}). To address this, we apply Gaussian refinement to the low-resolution reference points $\mathbf{P}$ \cite{zhang2020distribution}, resulting in corrected reference points $\mathbf{P}'=\{\rho'_1, \ldots, \rho'_s \}$. To specifically target foreground objects and avoid background segmentation, we compute the Hadamard product between the refined semantic mask $S'_m$ for class $p_m$ and the corrected reference points $\mathbf{P}'$ as \(\mathbf{Q}_m = S'_m \circ \mathbf{P}'\), where $\circ$ represents the Hadamard product, and $\mathbf{Q}_m \subseteq \mathbf{P}'$ denotes the set of resulting reference points for objects belonging to the label $p_m$, serving as a guide for identifying regions likely to contain the target objects, as illustrated in \cref{fig:reference_point}. This ensures the reference points $\mathbf{Q}_m$ to guide the segmentation to the target objects. These per-class reference points act as density maps (see ~\cref{fig:instvsacc}(b)), allowing for accurate object counting across varying densities. This automated selection can also be replaced by manual point or box annotations, making our model interactive for multiple user inputs.

\myparagraph{SAM mask generator:} By incorporating the reference object activation $F_\mathcal{P}$ and modifying the mask generation process, SAM's mask decoder can better focus on the reference object features. This additional contextual information from $F_\mathcal{P}$ helps the mask generator accurately distinguish and segment target objects. Since SAM's encoder requires an RGB image, we extract an RGB patch $I_m$ of the target object by multiplying the input image $I$ with the refined semantic mask $S'_m$ (see \cref{fig:refinement_depth}). The resultant mask $\mathbf{M}_m$ from SAM is obtained as \(\mathbf{M}_m = \text{SAM}(I_m, \mathbf{Q}_m)\), where $\mathbf{M}_m = \{M_{m_1}, \ldots, M_{m_n}\}$ is the set of individual object masks segmented by SAM. We count these masks to determine the total number of objects for class label $p_m$. This approach focuses on target objects without segmenting unrelated entities, enhancing efficiency and accuracy beyond the standard ``segment everything'' strategy. Finally, masks that are empty or cover an insignificantly small area are discarded, creating a refined subset $\mathbf{N}_m \subseteq \mathbf{M}_m$ containing only significant masks for the final object count. The cardinality of $\mathbf{N}_m$, $\texttt{card}(\mathbf{N}_m)$, denotes the final count of objects of class-label $m$ in image $I$.

\vspace{-3pt}

\begin{figure}[htbp]
\centering
    \includegraphics[width=0.85\columnwidth, keepaspectratio]{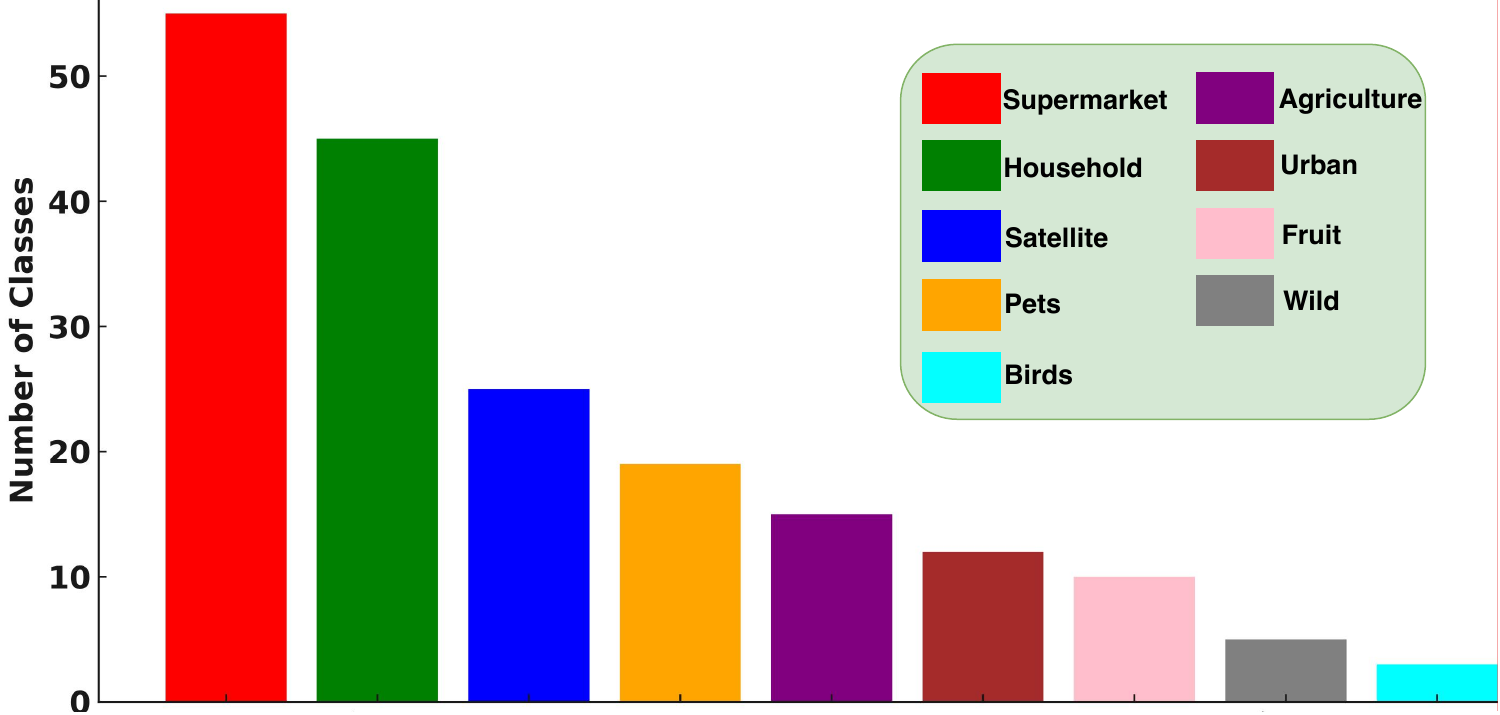}
    \caption{\textbf{OmniCount-191 statistics:} The number of categories per domain in long-tailed distribution format.}
    \label{fig:omni_stat}
\end{figure}

\vspace{-20pt}

\section{OmniCount-191 Dataset}
\label{sec:dataset}
To effectively evaluate OmniCount across open-vocabulary, supervised, and few-shot counting tasks, a dataset catering to a broad spectrum of visual categories and instances featuring various visual categories with multiple instances and classes per image is essential. The current datasets, primarily designed for object counting \cite{Ranjan2021CountEverything} focusing on singular object categories like humans and vehicles, fall short for multi-label object counting tasks. Despite the presence of multi-class datasets like MS COCO \cite{lin2014microsoft}, PASCAL VOC \cite{hoiem2009pascal}, and REC-8K \cite{Dai2024REC}, their utility is limited for counting due to the sparse nature of object appearance and fine-grained referencing. Addressing this gap, we created a new dataset with 30,230 images spanning 191 diverse categories, including kitchen utensils, office supplies, vehicles, and animals. This dataset features a wide range of object instance counts per image, ranging from 1 to 160, with an average of 10, bridging the existing gap and setting a benchmark for assessing counting models in varied scenarios.

\myparagraph{Dataset statistics:} The OmniCount-191 benchmark presents images with small, densely packed objects from multiple classes, reflecting real-world object counting scenarios. This dataset encompasses 30,230 images, with dimensions averaging 700 $\times$ 580 pixels. Each image contains an average of 10 objects, totalling 302,300, with individual images ranging from 1 to 160. We use the same annotation rules defined in existing counting datasets \cite{Ranjan2021CountEverything}. To ensure diversity, the dataset is split into training and testing sets, with no overlap in object categories -- 118 categories for training and 73 for testing, corresponding to a 60\%-40\% split. This results in 26,978 images for training and 3,252 for testing. Class splits are available for few and zero-shot settings for specific applications, detailed in the supplementary.

\vspace{-10pt}

\begin{table}[t]
\centering
\setlength{\tabcolsep}{1.5pt} 
\resizebox{\columnwidth}{!}{%
\begin{tabular}{@{}lccccc@{}} 
\hline
\multirow{2}{*}{Methods}                        & \multirow{2}{*}{Training} & \multicolumn{2}{c}{\textbf{PASCAL VOC}} & \multicolumn{2}{c}{\textbf{OmniCount-191}}  \\ 
\cline{3-6}
                                                &                           & mRMSE $\downarrow$            & mRMSE-nz $\downarrow$             & mRMSE $\downarrow$         & mRMSE-nz $\downarrow$          \\ 
\hline
ILC (\citeauthor{cholakkal2019object}) %
& \cmark                       & \underline{0.29}             & \underline{1.14}                 & {4.56}           & {9.39}               \\
CEOES (\citeauthor{Chattopadhyay2017CountEveryday}) %
& \cmark                       & {0.42}             & {1.65}                 & -             & -                 \\
Grounding-DINO (\citeauthor{liu2023grounding})%
& \xmark                        & 0.0066            & 0.05                & 1.29         & 3.27             \\
CLIPSeg (\citeauthor{luddecke2022image})%
& \xmark                        & 0.0091            & 0.08                & 1.54         & 4.28             \\
TFOC (\citeauthor{shi2024training})%
& \xmark                        & 0.0084            & 0.03                & 0.95         & 2.89             \\
\rowcolor[rgb]{0.53, 0.81, 0.92} {GrREC} (\citeauthor{Dai2024REC})%
& \cmark                        & -            & -                & \underline{0.50}         & \underline{1.87}             \\
\rowcolor[rgb]{0.753,0.753,0.753} \textbf{OmniCount} & \xmark                        & \textbf{0.0023} & \textbf{0.009}     & \textbf{0.70} & \textbf{2.00}     \\
\hline
\end{tabular}
}
\caption{\textbf{Performance comparison in multi-label object counting using text prompts.} Results on the PASCAL VOC and OmniCount-191 datasets. Methods requiring training are marked (\cmark). The best results are in \textbf{bold}, while the best scores among the learning-based methods are \underline{underlined}. Zero-shot models are marked \colorbox{skybluehighlight}{blue}.} 
\label{multi_text}
\vspace{-7pt}
\end{table}

\section{Experiments}
\label{sec:expt}
\myparagraph{Datasets:} For multi-label counting, we evaluate OmniCount on our proposed OmniCount-191 benchmark, specifically designed for multi-class scenarios. 
Additionally, following the detection and segmentation-based models \cite{Chattopadhyay2017CountEveryday} for multi-label counting, we compare OmniCount on the PASCAL VOC dataset \cite{hoiem2009pascal}, which includes 9963 images across 20 real-world classes, with 4952 designated for testing. For single-class counting, we used the test sets from FSC-147 \cite{Ranjan2021CountEverything} and CARPK \cite{hsieh2017drone}. Among them, FSC-147 includes 1190 images across 29 categories, while CARPK provides 1014 test images. FSC-147 and CARPK offer point and box annotations compatible with our model, whereas PASCAL VOC only provides box annotations.

\begin{table*}[!t]
\centering

\setlength{\tabcolsep}{5pt}
\resizebox{\linewidth}{!}{
\footnotesize
\begin{tabular}{lcccccccccc} 
\toprule
\textbf{Models} & \textbf{Training} & \textbf{Prompt} & \multicolumn{4}{c}{\textbf{FSC-147}} & \multicolumn{4}{c}{\textbf{CARPK}} \\ 
\cmidrule(lr){4-7} \cmidrule(lr){8-11}
& & & \textbf{MAE $\downarrow$} & \textbf{RMSE $\downarrow$} & \textbf{NAE $\downarrow$} & \textbf{SRE $\downarrow$} & \textbf{MAE $\downarrow$} & \textbf{RMSE $\downarrow$} & \textbf{NAE $\downarrow$} & \textbf{SRE $\downarrow$} \\ 
\midrule
CFOCNet+ \cite{yang2021class} & \cmark & box & 22.10 & 112.71 & - & - & - & - & - & - \\
GMN \cite{lu2019class} & \cmark & box & 26.52 & 124.57 & - & -7.48 & 9.90 & - & - & - \\
BMNet+ \cite{shi2022represent} & \cmark & box & \underline{14.62} & {91.83} & \underline{0.25} & \underline{2.74} & \underline{5.76} & \underline{7.83} & - & - \\
Vanilla SAM \cite{kirillov2023segment} & \xmark & N.A. & 42.48 & 137.50 & 1.14 & 8.13 & 16.97 & 20.57 & 0.70 & 5.30 \\
\rowcolor[rgb]{0.53, 0.81, 0.92} PSeCo \cite{huang2024point} & \cmark & N.A. & 16.58 & 129.77 & - & - & - & - & - & - \\
TFOC \cite{shi2024training} & \xmark & box & 19.95 & 132.16 & 0.29 & 3.80 & 10.97 & 14.24 & 0.48 & 3.70 \\
\rowcolor[rgb]{0.753,0.753,0.753}
\textbf{OmniCount} & \xmark & box & \textbf{18.63} & \textbf{112.98} & \textbf{0.14} & \textbf{2.99} & \textbf{9.92} & \textbf{12.15} & \textbf{0.23} & \textbf{2.11} \\ 
TFOC \cite{shi2024training} & \xmark & point & 20.10 & 132.83 & 0.30 & 3.87 & 11.01 & 14.34 & 0.51 & 3.89 \\
\rowcolor[rgb]{0.753,0.753,0.753}
\textbf{OmniCount} & \xmark & point & \textbf{19.24} & \textbf{115.27} & \textbf{0.25} & \textbf{3.21} & \textbf{10.66} & \textbf{13.15} & \textbf{0.31} & \textbf{2.45}  \\
\rowcolor[rgb]{0.53, 0.81, 0.92} 
ZSOC \cite{xu2023zero} & \cmark & text & {22.09} & {115.17} & {0.34} & {3.74} & {-} & {-} & - & - \\
TFOC \cite{shi2024training} & \xmark & text & 24.79 & 137.15 & 0.37 & 4.52 & - & - & - & - \\
\rowcolor[rgb]{0.53, 0.81, 0.92} 
GrREC \cite{Dai2024REC} & \cmark & text & \underline{10.12} & \underline{107.19} & - & - & - & - & - & - \\
\rowcolor[rgb]{0.753,0.753,0.753}
\textbf{OmniCount} & \xmark & text & \textbf{21.46} & \textbf{133.28} & \textbf{0.32} & \textbf{0.39} & - & - & - & - \\
\bottomrule
\end{tabular}}
\caption{\textbf{Results in single-label object counting} 
setting using text, point, and box prompts. The \textbf{bold} denotes the best among training-free methods, while the \underline{underlined} font is the best among learning-based methods. Zero-shot models are marked \colorbox{skybluehighlight}{blue}.}
\vspace{-16pt}
\label{tab:single_label}
\end{table*}

\begin{figure}[!t]
\centering
    \includegraphics[width=\linewidth]{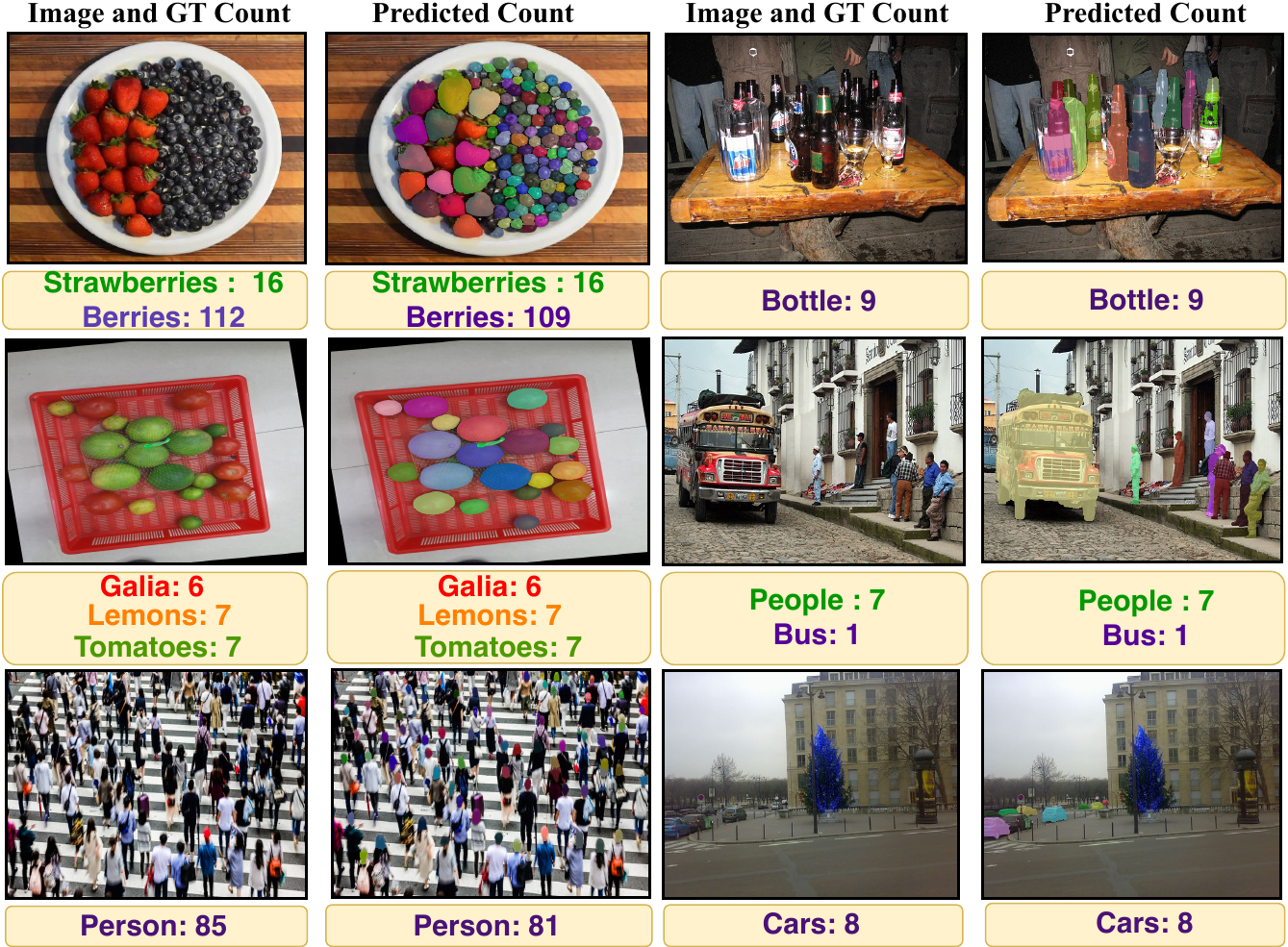}
    \caption{\textbf{Qualitative Results using OmniCount}: OmniCount-191 (left), PASCAL VOC (right).}
    \label{viz}
\end{figure}
\subsection{Multi-label object counting}
\myparagraph{Competitors:} For a fair comparison with state-of-the-art methods, we adapt the single-label object counting model TFOC \cite{shi2024training} for multi-label counting by running it for each class in an image to obtain multi-label counts. We also replicate and evaluate the performance of ILC \cite{cholakkal2019object} and GrREC \cite{Dai2024REC} for a direct comparison, though replicating CEOES \cite{Chattopadhyay2017CountEveryday} was limited by its Lua implementation. We additionally employ open-vocabulary object detection (Grounding-DINO \cite{liu2023grounding}) and semantic segmentation (CLIPSeg \cite{luddecke2022image}) baselines for multi-label counting. The Grounding-DINO baseline counts objects by enumerating detected bounding boxes per category, while the CLIPSeg baseline uses a ViT encoder and spectral clustering to estimate category counts by identifying connected components.

\begin{figure}[!t]
    \centering
    \includegraphics[width=\linewidth]{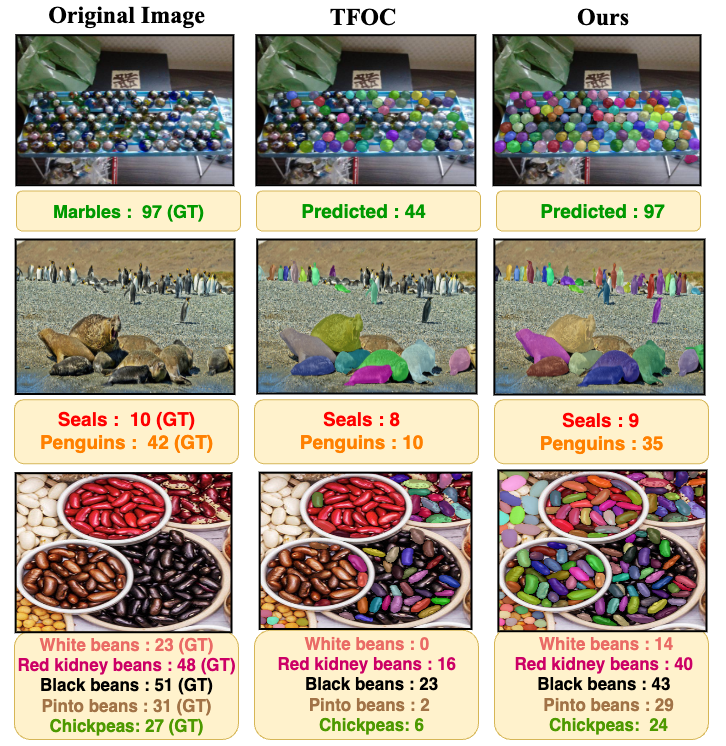}
    \caption{\textbf{Qualitative comparisons} with TFOC on the OmniCount-191 dataset.}
    \label{vizcomp}
 \vspace{-10pt}
\end{figure}

\myparagraph{Results:} In \cref{multi_text}, we compare OmniCount with existing multi-label counting methods, demonstrating its strong performance, especially as a training-free model. Although the recently introduced GrREC, a training-based method, achieves slightly better scores, OmniCount remains highly competitive despite not being trained on seen classes. This highlights the benefits of our open-vocabulary approach, which uses geometric priors to accurately count multiple categories in a single pass -- unlike traditional models that struggle with occlusions and require separate passes for each category.
Notably, our SAM-based OmniCount surpasses the CLIPSeg and Grounding-DINO baselines, confirming SAM's effectiveness for counting tasks. Qualitative results in \cref{viz} show OmniCount's performance on OmniCount-191 and PASCAL VOC, while \cref{vizcomp} compares it with TFOC on OmniCount-191. These results highlight OmniCount's robustness in counting objects of various sizes, from large singular items like seals and buses to medium-sized objects (\eg bottles, cars etc.) and small, non-atomic entities like pulses and berries. Further analysis using ground-truth bounding box and point annotations is provided in the supplementary material.

\subsection{Single-label counting}
\myparagraph{Competitors:} We report the performance of training-based methods like CFOCNet+ \cite{yang2021class}, GMN \cite{lu2019class}, BMNet \cite{shi2022represent}, ZSOC \cite{xu2023zero}, PSeCo \cite{huang2024point}, GrREC \cite{Dai2024REC}, as well as training-free approaches like TFOC \cite{shi2024training}. We have also adopted a SAM-based baseline for a fair comparison, reporting Vanilla SAM \cite{kirillov2023segment} counting results by processing entire images with a uniform point layout.

\myparagraph{Results:} We rigorously compare our model's performance in a single-label context utilizing text, box, and point prompts, as shown in \cref{tab:single_label}. Like multi-label counting, OmniCount consistently outperforms major training-based models, and all training-free models across all the text/box/point prompt modalities across four key metrics demonstrate its robustness and efficiency in object counting tasks. This also illustrates that merely using SAM as a counting model is inferior, even in single-class counting, highlighting the importance of different priors. More results and insights on other OmniCount-191 tasks, such as VQA, have been provided in the supplementary material.

\begin{figure}[!t]
    \centering
    \includegraphics[width=\linewidth]{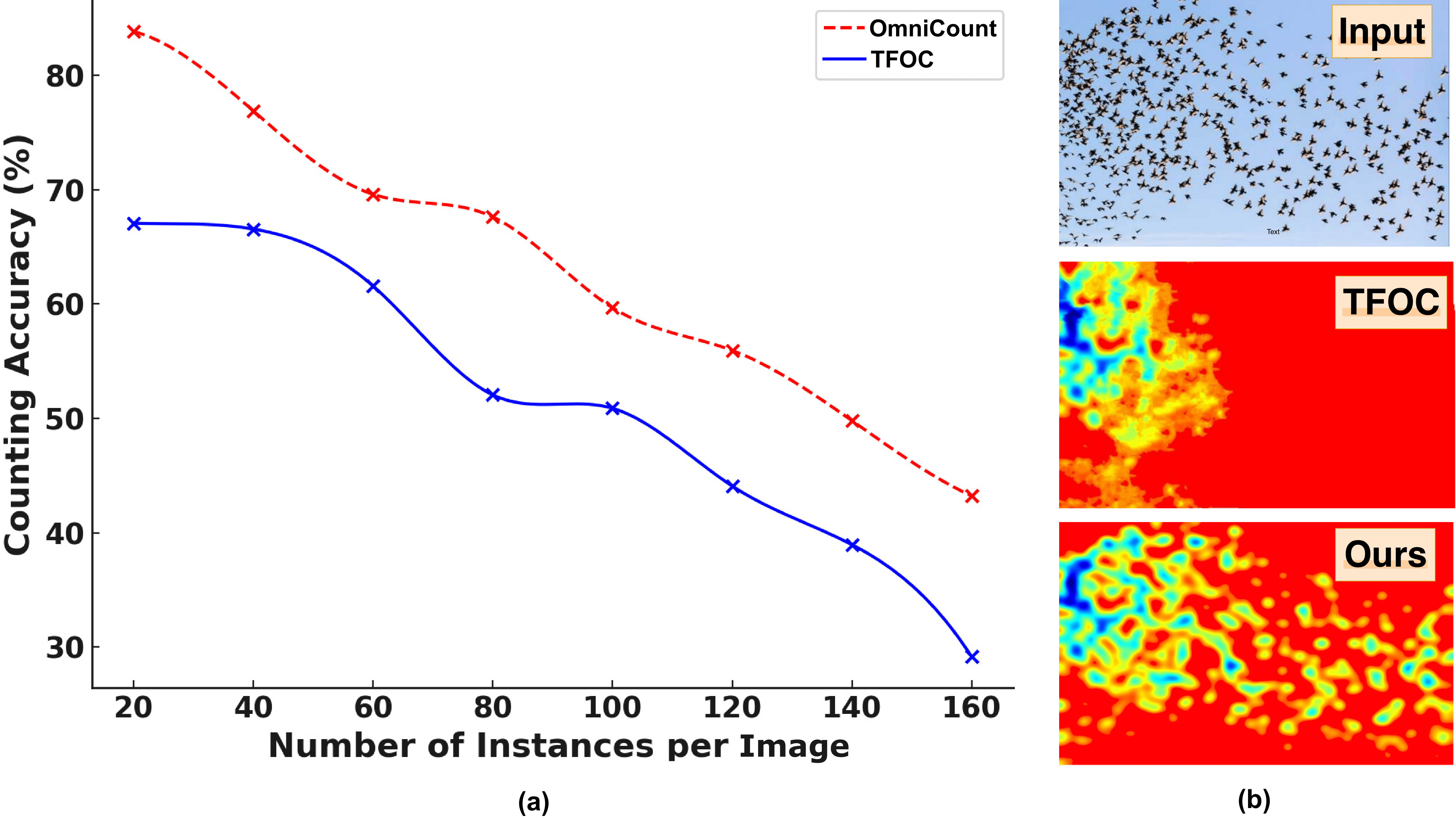}
    \caption{Performance comparison in dense scenes. (a) Counting accuracy vs number of instances per image (b) Counting heatmap in varying depth image}
    \label{fig:instvsacc}
\end{figure}

\subsection{Further analysis}
\myparagraph{Impact of depth refinement:} OmniCount leverages semantic (SP) and geometric priors (GP) to improve SAM's segmentation performance, making it suitable as an object counter. We assess the impact of SP and GP on SAM's object counting performance using the OmniCount-191 dataset, as shown in \cref{table:ablations}. The best results (rows 2-4) indicate that without GP, OmniCount has 56/8\% higher error rate in m-RMSE/m-RMSE-nz metrics, suggesting that SAM over-segments and over-counts when it lacks structural/geometric information for occluded objects.

\begin{table}
\centering
\footnotesize
\begin{tabular}{ccc|cc}
\hline
\textbf{SP}                         & \textbf{GP}                         & \textbf{RP}                         & \textbf{m-RMSE} $\downarrow$                     & \textbf{m-RMSE-nz} $\downarrow$                \\ \hline
\cmark                              & \xmark                             & \xmark                             & 7.02                                & 5.89                                \\
\cmark                              & \xmark                             & \cmark                             & 1.62                                & 2.17                                \\
\cmark                              & \cmark                             & \xmark                             & 2.12                                & 2.54                                \\
\rowcolor[HTML]{C0C0C0} 
{\color[HTML]{000000} \textbf{\cmark}} & {\color[HTML]{000000} \textbf{\cmark}} & {\color[HTML]{000000} \textbf{\cmark}} & {\color[HTML]{000000} \textbf{0.70}} & {\color[HTML]{000000} \textbf{2.00}} \\ \hline
\end{tabular}
\caption{Ablation of Semantic Prior (SP), Geometric Prior (GP) and Reference Point (RP) on OmniCount-191 dataset.} 
\label{table:ablations}
\end{table}

\myparagraph{Importance of reference points:}
OmniCount employs reference point (RP) selection using the feature activation $F_\mathcal{P}$ from semantic priors, feeding the selected RPs into SAM for segmentation and counting. In this experiment, we have evaluated the impact of RP selection versus SAM's default ``everything mode'' on object counting using the OmniCount-191 dataset, as shown in \cref{table:ablations}. The best results (rows 3-4) indicate that SAM's ``everything mode'' with uniform point selection leads to overcounting, with a 67/21\% increase in error rates, showing the effectiveness of our reference point selection step.

\myparagraph{Counting ability in dense scenarios:} In \cref{fig:instvsacc}, we compare the counting performance of OmniCount and TFOC in dense scenes from OmniCount-191. As shown in \cref{fig:instvsacc}(a), both models experience a decline in performance as the number of instances per image increases. This deterioration is more pronounced in TFOC due to its reliance on segmentation methods, which struggle with occlusions and dense object distributions. This also justifies why object counting approaches do not work well on crowd counting \cite{pelhan2024dave}. Adding depth priors helps mitigate these errors, as reflected in OmniCount's improved performance. Similarly, in \cref{fig:instvsacc}(b), while TFOC can count birds close to the camera, our method can identify them at varying depths, even ones in the sky with infinite depth.

\vspace{-0.1in}
\section{Conclusion}
\label{sec:concl}

We introduced OmniCount, a novel open-vocabulary, multi-label counting model capable of processing multiple categories in a single pass, integrating semantic and geometric insights without requiring training. Surpassing traditional, category-specific models limited by dataset constraints, OmniCount utilizes pre-trained foundation models for semantic segmentation and depth estimation to address occlusions and achieve precise object segmentation and counting. To fill the void of a dedicated multi-label counting dataset, we developed OmniCount-191, featuring 30,230 images across 191 categories. OmniCount's efficacy, tested on existing benchmarks and our OmniCount-191 in various settings, showcases its superior performance, efficiency, and scalability, emphasizing its readiness for real-world applications and establishing multi-label counting as a practical, feasible tool.

\newpage


\section{Dataset Description}
\label{data_des}

\subsection{Image Collection}
\begin{figure}[!htb]
  \centering
   \includegraphics[width=\columnwidth, keepaspectratio]{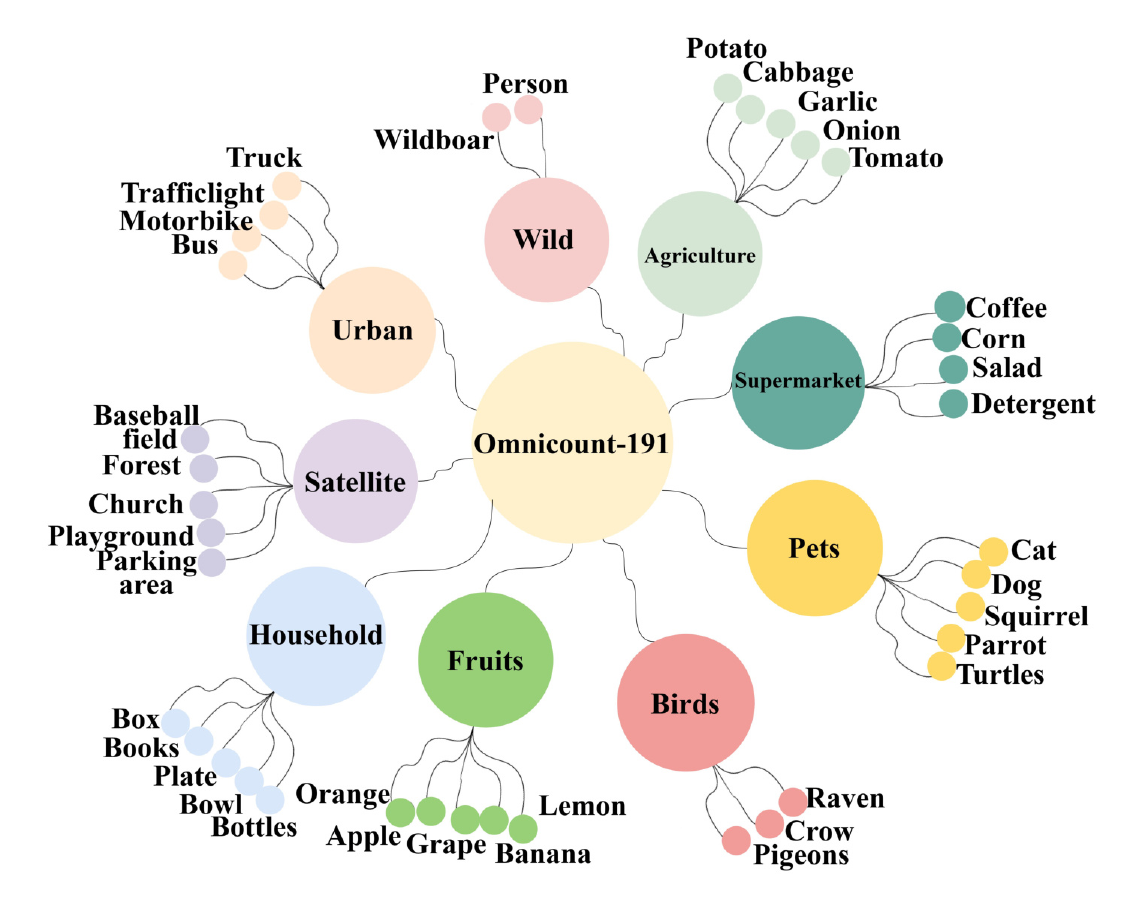}
  \caption{\textbf{A concise overview of the OmniCount-191 dataset}: This dataset features images across nine diverse domains, encompassing a wide range of object densities, shapes, and sizes, making it perfectly suited for object counting tasks. The figure shows the most frequent object categories present per domain.}
  \label{fig:dataset_stat}
\end{figure}

Our dataset OmniCount-191, consisting of 30,230 images, was carefully collected to ensure utility and quality. It was curated from a broad collection of candidate images identified through keyword searches across 191 real-world object categories, as shown in \cref{fig:dataset_stat}. The selection was refined through a detailed manual review, adhering to stringent criteria: (1) \textbf{Object instances:} Each image must contain at least \emph{five} object instances, aiming to challenge object enumeration in complex scenarios; (2) \textbf{Image quality:} High-resolution images were selected to ensure clear object identification and counting; (3) \textbf{Severe occlusion:} We excluded images with significant occlusion to maintain accuracy in object counting; (4) \textbf{Object dimensions:} Images with objects too small or too distant for accurate counting or annotation were removed, ensuring all objects are adequately sized for analysis.
This selection process crafted a dataset poised to advance object counting algorithm development and testing, tailored for real-world applicability. We follow the annotation criteria similar to single-object counting benchmarks like FSC-147 \cite{Ranjan2021CountEverything}

\subsection{Dataset Curation} The data collection process for OmniCount-191 involved a team of 13 members who manually curated images from the web, released under Creative Commons (CC) licenses. The images were sourced using relevant keywords such as ``Aerial Images'', ``Supermarket Shelf'', ``Household Fruits'', and ``Many Birds and Animals''. Initially, 40,000 images were considered, from which 30,230 images were selected based on the guidelines outlined in Section 4. The selected images were annotated using the Labelbox \cite{sharma2019labelbox} annotation platform.
As shown in \cref{tab:dataset_comp}, most existing object counting datasets have been designed for specific object categories \cite{idrees2013multi, idrees2018composition, zhang2016single, wang2020nwpu, sindagi2019pushing, Sindagi2022JHUCrowd, hsieh2017drone}. The FSC-147 \cite{Ranjan2021CountEverything} dataset was the first to include multiple object categories, however, FSC-147 does not contain annotations of multiple categories in a single image. So, there was still a lack of a comprehensive dataset containing multi-label annotations per image and annotations for tasks like Visual Question Answering (VQA) for counting. OmniCount-191 aims to fill this gap by providing a diverse and comprehensive dataset with multi-label annotations and support for various counting-related computer vision tasks.

\begin{table*}[!t]
\centering
\setlength{\extrarowheight}{1pt}
\addtolength{\extrarowheight}{\aboverulesep}
\addtolength{\extrarowheight}{\belowrulesep}
\setlength{\aboverulesep}{0pt}
\setlength{\belowrulesep}{0pt}
\begin{tabular}{c|ccc|cccc}
\hline
                                    &                                                                                       &                                                                                           &                                                                                        & \multicolumn{4}{c}{\textbf{Annotation}}                         \\ \cline{5-8} 
\multirow{-2}{*}{\textbf{Datasets}} & \multirow{-2}{*}{\textbf{\begin{tabular}[c]{@{}c@{}}Annotated\\ Images\end{tabular}}} & \multirow{-2}{*}{\textbf{\begin{tabular}[c]{@{}c@{}}Number of\\ Categories\end{tabular}}} & \multirow{-2}{*}{\textbf{\begin{tabular}[c]{@{}c@{}}Labels per\\ Image\end{tabular}}} & \textbf{Point} & \textbf{Box} & \textbf{VQA} & \textbf{Caption} \\ \hline
UCF CC 50 \cite{idrees2013multi}                           & 50                                                                                    & 1                                                                                         & Single                                                                                 & \cmark            & \xmark           & \xmark           & \xmark               \\
Shanghaitech  \cite{zhang2016single}                      & 1198                                                                                  & 1                                                                                         & Single                                                                                 & \cmark            & \xmark           & \xmark           & \xmark               \\
UCF QNRF  \cite{idrees2018composition}                          & 1535                                                                                  & 1                                                                                         & Single                                                                                 & \cmark            & \xmark           & \xmark           & \xmark               \\
NWPU \cite{wang2020nwpu}                                & 5109                                                                                  & 1                                                                                         & Single                                                                                 & \cmark            & \xmark           & \xmark           & \xmark               \\
JHU Crowd \cite{sindagi2019pushing}                           & 4372                                                                                  & 1                                                                                         & Single                                                                                 & \cmark            & \cmark          & \xmark           & \xmark               \\
CARPK \cite{hsieh2017drone}                              & 1148                                                                                  & 1                                                                                         & Single                                                                                 & \cmark            & \cmark          & \xmark           & \xmark               \\
PASCAL VOC \cite{hoiem2009pascal}                         & 1449                                                                                  & 20                                                                                        & Multi                                                                                  & \xmark             & \cmark          & \xmark           & \xmark               \\
FSC-147 \cite{Ranjan2021CountEverything}                            & 6135                                                                                  & 147                                                                                       & Single                                                                                 & \cmark            & \cmark          & \xmark           & \xmark               \\
{REC-8K \cite{Dai2024REC}}
                            & 8011                                                                                  & -                                                                                       & Multi                                                                                 & \cmark            & \cmark          & \xmark           & \cmark               \\
\rowcolor[HTML]{C0C0C0} 
\textbf{OmniCount-191}              & \textbf{30,230}                                                                       & \textbf{191}                                                                              & \textbf{Multi}                                                                         & \textbf{\cmark}   & \textbf{\cmark} & \textbf{\cmark} & \textbf{\cmark}     \\ \hline
\end{tabular}
\caption{Comparison with existing benchmark datasets}
\label{tab:dataset_comp}
\end{table*}

\section{Implementation Details and Metrics}
\label{implementation}
In our experiments, we employ ``ViT-Large'' for SAN \cite{xu2023side} and ``ViT-Base'' for SAM \cite{kirillov2023segment} models. For the k-nearest neighbor, we use a 10-pixel search window and set a depth threshold $\tau=0.3$ to accommodate the depth variance of objects with curved edges. The values of $K$ and $C$ are respectively set as $16$ and $256$. For our prior-guided mask generation, we select the local maxima in SAN's heatmap \cite{xu2023side}, then refine them using Gaussian refinement with $\sigma = 0.4$ and $\omega = 4$. Finally, we input them as reference object points into SAM for mask generation and counting. Additionally, we compare box and point prompts with traditional counting methods \cite{shi2024training}. For the former, bounding boxes from PASCAL VOC \cite{hoiem2009pascal} and OmniCount-191 datasets serve as prompts for SAM. For datasets having no point annotation, we calculate the centroid of each bounding box and use its coordinates as prompts. We will release the code upon acceptance.

\myparagraph{Evaluation metrics:} For evaluating our model's performance in single-class object counting, we employ four key metrics in line with leading benchmarks \cite{shi2024training, Ranjan2022ExemplarFree, Ranjan2021CountEverything}: Mean Average Error (MAE) for standard accuracy assessment, Normalized Mean Average Error (NMAE) for a more intuitive understanding of errors, along with Normalized Relative Error (NAE) and Squared Relative Error (SRE) for comprehensive error analysis. In multi-label counting, we use mean-RMSE ({errors averaged across all categories, denoted by} mRMSE) and nonzero-RMSE ({errors averaged over all ground-truth instances with non-zero counts}, denoted by mRMSE-nz) to assess the model's precision across various object categories, following the prior works \cite{cholakkal2019object, Cholakkal2022PartialSupervision, Chattopadhyay2017CountEveryday}.


\section{Further analyses}
\label{ex_ablation}
\subsection{Counting efficiency:} 
\begin{figure}[!htb]
  \centering
   \includegraphics[width=\columnwidth, keepaspectratio]{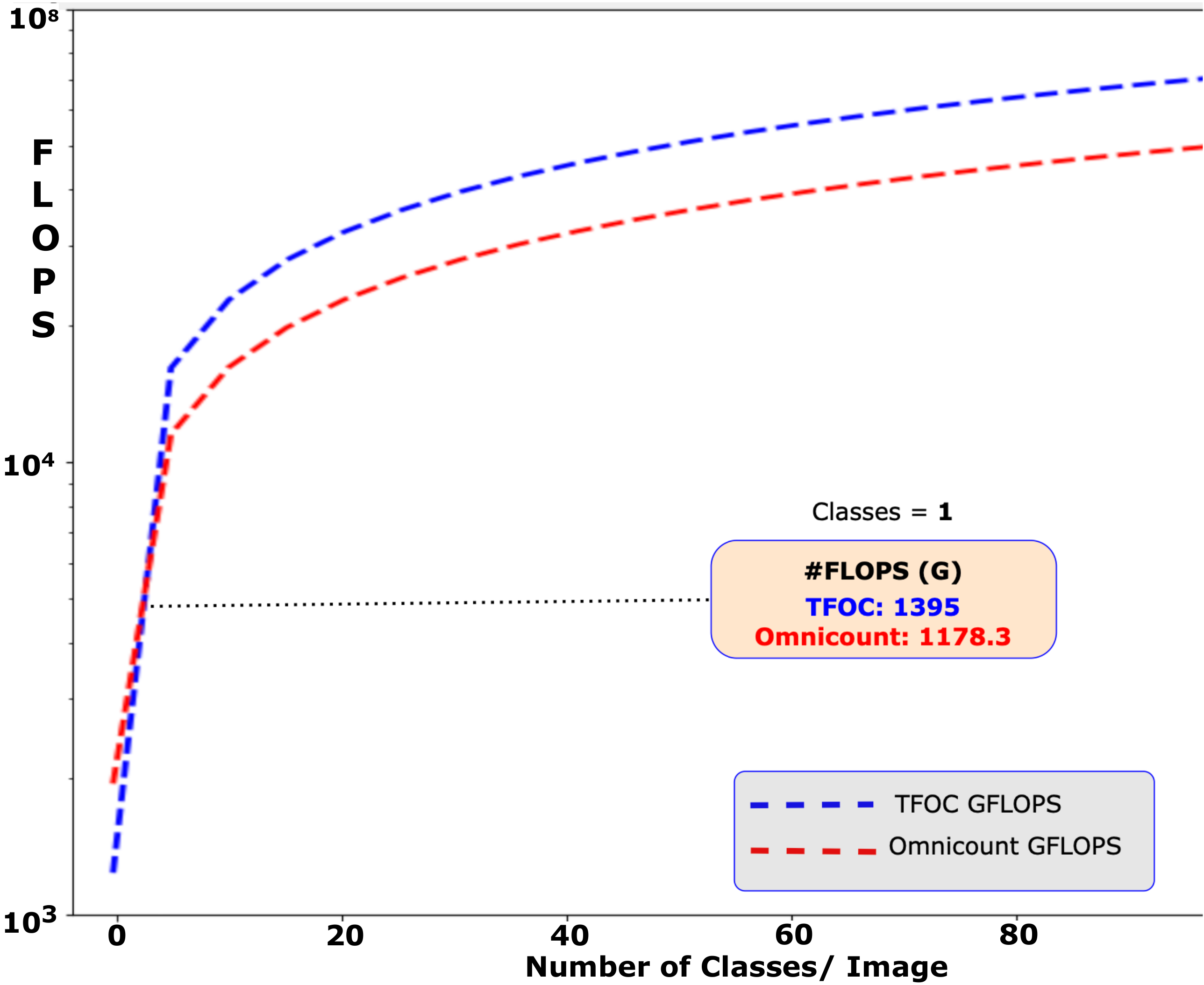}
  \caption{Scalability versus efficiency plots comparing TFOC and OmniCount.}
  \label{fig:efficiency}
\end{figure}

{When comparing the scalability and efficiency of the benchmark training-free counting model TFOC with our OmniCount, we evaluated their computational complexity. The graph in \cref{fig:efficiency} illustrates that TFOC's GFLOPS increase exponentially as the number of object categories increases. Conversely, OmniCount demonstrates linear growth due to having a class-specific SAM \cite{kirillov2023segment} component, which grows linearly with classes while other components remain constant. This indicates superior scalability and efficiency in diverse object scenarios.}

\subsection{Depth Estimation Module}

{We evaluated the performance of depth estimation models, comparing the diffusion-based Marigold \cite{ke2023repurposing} with the non-diffusion-based MiDAS \cite{birkl2023midas}. As shown in \cref{table:depth_estimation}, Marigold consistently outperforms MiDAS. This superior performance can be attributed to Marigold’s use of pre-trained generative diffusion models, which enable it to tap into a vast repository of prior knowledge. This allows Marigold to deliver more accurate and efficient monocular depth estimations, particularly in complex scenes where traditional methods might struggle. Integrating diffusion models provides Marigold with a distinct advantage in refining depth estimates, leading to more precise and reliable outputs.}

\begin{table}[ht]
\centering
\begin{minipage}{0.5\textwidth}
\centering
\resizebox{\linewidth}{!}{%
\begin{tabular}{c|cc}
\hline
Model & m-RMSE $\downarrow$ & m-RMSE-nz $\downarrow$ \\
\hline
MIDAS \cite{birkl2023midas} & 0.91 & 2.07 \\
\rowcolor{gray!30}
\textbf{Marigold} \cite{ke2023repurposing} & \textbf{0.70}  & \textbf{2.00} \\
\hline
\end{tabular}}
\caption{Choice of depth models}
\label{table:depth_estimation}
\end{minipage}
\hfill
\begin{minipage}{0.5\textwidth}
\centering
\resizebox{\linewidth}{!}{%
\begin{tabular}{c|cc}
\hline
Model & mRMSE $\downarrow$ & mRMSE-nz $\downarrow$ \\
\hline
SimSeg \cite{luddecke2022image} & 2.99 & 5.22 \\
OvSeg \cite{liang2023open} & 1.72 & 3.10 \\
\rowcolor{gray!30}
\textbf{SAN} \cite{xu2023side} & \textbf{0.70} & \textbf{2.00} \\
\hline
\end{tabular}}
\caption{Choice of semantic priors}
\label{table:semantic_estimation}
\end{minipage}
\end{table}


\begin{table}[ht]
\centering
\begin{minipage}{0.5\linewidth}
\centering
\resizebox{0.8\linewidth}{!}{%
\begin{tabular}{cc} 
\toprule
{Model} & {mAP} $\uparrow$\\ 
\midrule
TFOC & 23.27 \\
\rowcolor{gray!30}
\textbf{Omnicount} & \textbf{39.32} \\
\bottomrule
\end{tabular}}
\caption{Localization performance}
\label{table:loc}
\end{minipage}%
\hfill
\begin{minipage}{0.5\linewidth}
\centering
\resizebox{\linewidth}{!}{%
\begin{tabular}{cc} 
\toprule
{Model} & {mRMSE} $\downarrow$\\ 
\midrule
Ours (no SAM) & 0.98 \\
\rowcolor{gray!30}
Ours (with SAM) & 0.70 \\
\bottomrule
\end{tabular}}
\caption{Directly using reference points as count}
\label{table:point}
\end{minipage}
\end{table}

\subsection{Semantic Estimation Module}
In \cref{table:semantic_estimation}, we evaluated various open-vocabulary semantic segmentation modules, including SimSeg \cite{xu2022simple}, OVSeg \cite{liang2023open}, and SAN \cite{xu2023side} for generating semantic priors, with SAN identified as the top performer due to its integration with CLIP. SAN's utilization of a side network augments CLIP's capabilities and promotes efficient feature reuse using a lightweight architecture. Its end-to-end training methodology seamlessly aligns with CLIP, improving mask proposal accuracy and outperforming other models in semantic segmentation efficiency with fewer parameters.

\subsection{Localisation Performance}


{Here we aimed to evaluate the object localization accuracy and compared OmniCount against the baseline model TFOC. As observed in \cref{table:loc}, OmniCount demonstrates a significant improvement in localization accuracy, achieving an mAP score of 39.32, compared to TFOC’s 23.27. This result highlights the effectiveness of OmniCount in precisely identifying object locations besides accurate counting.}

\subsection{Using points directly for counting}
We provide the performance of our model by directly counting the reference points generated following Gaussian refinement without passing them as prompts to SAM in \cref{table:point}. We can see using SAM significantly improves OmniCount's performance.

\subsection{Overlap of pretraining categories of SAN vs OmniCount-191}
In \cref{tab:pretrain}, we evaluate semantic robustness on classes like satellites, pets, and household items, absent in COCO and COCO-Stuff, which SAN was fine-tuned on. The results indicate that our model generalizes better across diverse, non-overlapping categories than TFOC, showing less reliance on SAN’s pre-training.

\begin{table}[!t]
\centering
\resizebox{0.5\linewidth}{!}{%
\begin{tabular}{cc} 
\toprule
{Model} & {mRMSE} $\downarrow$\\ 
\midrule
TFOC & 0.83 \\
\rowcolor{gray!30}
Omnicount & 0.56 \\
\bottomrule
\end{tabular}
}
\caption{Effect of semantic pretraining}
\label{tab:pretrain}
\end{table}


\section{Omnicount performance using box and point prompts}
\label{box-pt}
In \cref{multi_box_point}, we report the performance of our model against TFOC \cite{shi2024training} using both box and point prompts. This evaluation bypasses our reference point selection module, leveraging the ground truth bounding box and point annotations from the datasets as prompts for SAM. The table shows that our model outperforms the baseline under this setting. Remarkably, the performance disparity between text-prompt (Table 1 in main paper) and box-point-prompt settings for OmniCount is minimal, underscoring the robustness of our reference point selection strategy.

\begin{table}[!htbp]
\centering
\resizebox{\columnwidth}{!}{
\begin{tabular}{@{}cccccc@{}}
\hline
\multirow{2}{*}{Prompt} & \multirow{2}{*}{Methods} & \multicolumn{2}{c}{\textbf{Pascal-VOC}} & \multicolumn{2}{c}{\textbf{OmniCount-191}} \\
\cline{3-6}
& & mRMSE $\downarrow$ & mRMSE-nz $\downarrow$& mRMSE $\downarrow$& mRMSE-nz $\downarrow$\\
\hline
\multirow{2}{*}{Box} & TFOC & 0.0067 & 0.018 & 0.91 & 2.78 \\
& \cellcolor[rgb]{0.753,0.753,0.753}\textbf{OmniCount} & \cellcolor[rgb]{0.753,0.753,0.753}\textbf{0.00185} & \cellcolor[rgb]{0.753,0.753,0.753}\textbf{0.00790} & \cellcolor[rgb]{0.753,0.753,0.753}\textbf{0.814} & \cellcolor[rgb]{0.753,0.753,0.753}\textbf{2.24} \\
\hline
\multirow{2}{*}{Point} & TFOC & 0.0072 & 0.025 & 0.917 & 2.85 \\
& \cellcolor[rgb]{0.753,0.753,0.753}\textbf{OmniCount} & \cellcolor[rgb]{0.753,0.753,0.753}\textbf{0.00190} & \cellcolor[rgb]{0.753,0.753,0.753}\textbf{0.00821} & \cellcolor[rgb]{0.753,0.753,0.753}\textbf{0.83} & \cellcolor[rgb]{0.753,0.753,0.753}\textbf{2.29} \\
\hline
\end{tabular}}
\caption{Performance of various approaches in multi-label object counting setting using box and point prompts. The best results are in \textbf{bold}. OmniCount demonstrates better performance even against training-based benchmarks.}
\label{multi_box_point}
\end{table}

\section{Zero-shot and Few-shot Splits for OmniCount-191}
\label{splits}
We have prepared dedicated splits within the OmniCount-191 dataset to facilitate the assessment of object-counting models under zero-shot and few-shot learning conditions.

\subsection{Zero-shot Split}

For the zero-shot split, we partition the dataset into training ($\mathcal{D}_\text{train}$) and testing ($\mathcal{D}_\text{test}$) sets, ensuring no overlap in categories between them ($\mathcal{D}_\text{train} \cap \mathcal{D}_\text{test} = \phi$). Specifically, the dataset's 191 object categories are divided following a $60\%-40\%$ ratio, allocating 118 categories to the training set and 73 to the testing set, aligning with the guidelines outlined in Sec. 4 of the main paper. To further guarantee the separation between training and testing conditions, images originating from distinct domains such as satellite imagery, birds, and urban landscapes are exclusively reserved for the testing set, while the remainder are allocated to the training set. This careful categorization ensures a rigorous evaluation framework for exploring the capabilities of object-counting models in zero-shot scenarios.

\subsection{Few-shot split}

For the few-shot learning evaluation, we structure the dataset into subsets designed to simulate scenarios where only a limited number of examples are available for model training. This division creates a practical setting to test the adaptability and efficiency of object-counting models when faced with minimal data.

In the few-shot split of OmniCount-191, we allocate a subset of images from each of the 191 categories, ensuring a balanced representation across different domains. Specifically, we designate a small number of images (ranging from 1 to 5) per category for the training set ($\mathcal{D}_\text{train}^{\text{few-shot}}$), while the remainder of the dataset forms the testing set ($\mathcal{D}_\text{test}^{\text{few-shot}}$). This setup aligns with the few-shot learning paradigm, where models must learn to generalize from a few examples.
While creating the few-shot split, we adopt the following detailed strategy:

\begin{enumerate}
    \item \textbf{Domain Selection}: Each of the nine domains represented in OmniCount-191 -- supermarket, fruits, urban, satellite, wild, household, pets, birds, and agriculture—is included in the few-shot learning evaluation. This diversity ensures that the models are tested across a wide range of contexts, from densely populated urban images to the varied species in wild and birds domains.

    \item \textbf{Class Allocation}: From each domain, a proportionate number of classes are selected for the few-shot training subset. For instance, if a domain like fruits has a high representation in the dataset, more classes from this domain are chosen for the few-shot split compared to a less represented domain. This allocation respects the dataset's inherent diversity while adhering to the few-shot learning constraints.

    \item \textbf{Image Selection}: For the selected classes in each domain, a predetermined number of images (e.g., 1-shot, 3-shot, or 5-shot) are randomly chosen for $\mathcal{D}_\text{train}^{\text{few-shot}}$. The selection process is carefully randomized to ensure that the few-shot training set is representative of the variability within each class and domain.

    \item \textbf{Testing Set Composition}: The remainder of the dataset, which includes the non-selected images from the few-shot classes and all images from the classes not designated for few-shot training, comprises the testing set ($\mathcal{D}_\text{test}^{\text{few-shot}}$). This ensures a robust testing environment where the model's ability to generalize from limited information can be accurately assessed.
\end{enumerate}

By incorporating classes from each of the nine domains, this few-shot split offers a comprehensive challenge to object-counting models, emphasizing the importance of adaptability and the efficient use of sparse data.

\section{Visual Question Answering on OmniCount-191}
\label{vqa}
\begin{figure*}[htbp]
\centering
    \includegraphics[width=\linewidth]{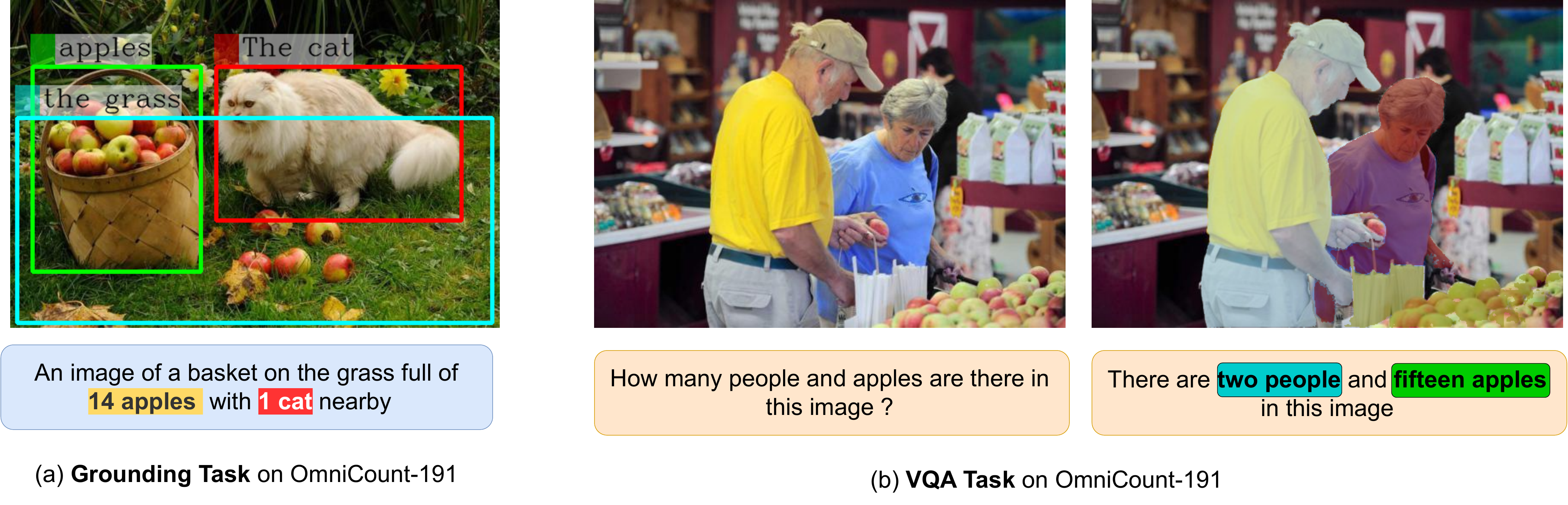}
    \caption{Sample output demonstrating visual question answering and visual grounding capabilities within OmniCount-191.}
    \label{fig:ground}
\end{figure*}

\begin{table*}[htbp]
\centering
\caption{Performance comparison on our OmniCount-191 and VQA-v2 \cite{Antol2015VQA}. VQA-v2 is referenced as a widely recognized benchmark in VQA research, included here for broader context and performance comparison.}
\label{tab:performance_comparison}

\begin{minipage}{0.3\linewidth}
\centering
\begin{tabular}{ccc} 
\toprule
\multirow{2}{*}{Method} & \multicolumn{2}{c}{Accuracy (\%) $\uparrow$}  \\
                        & OmniCount-191 & VQA-v2             \\ 
\midrule
ViLT \cite{kim2021vilt}                   & 14.45          & 71.26              \\
BLIP \cite{li2022blip}                    & 36.78         & 52.3               \\
\bottomrule
\end{tabular}
\end{minipage}%
\hfill
\begin{minipage}{0.5\linewidth}
\centering
\begin{tabular}{cc} 
\toprule
\multirow{2}{*}{Method} & mRMSE $\downarrow$          \\
                        & OmniCount-191  \\ 
\midrule
ViLT \cite{kim2021vilt}                   & 1.64          \\
BLIP \cite{li2022blip}                   & 0.31          \\
\bottomrule
\end{tabular}
\end{minipage}

\end{table*}

In OmniCount-191, we extend beyond traditional annotations to include Visual Question Answering (VQA) tasks focused on counting, enhancing the dataset's applicability across various domains such as image retrieval \cite{kafle2017visual, feng2023vqa4cir}, visual grounding \cite{chen2022grounding}, and more. This innovative integration effectively marries object counting with complex scene comprehension, expanding the dataset's utility.

For each image, we have crafted question and answer pairs (see \cref{fig:ground}), assessing object counting models through specific queries like ``How many people and apples are there in the image?'' and evaluating responses such as ``There are two people and fifteen apples''. The evaluation of these object counting models on OmniCount-191 (\cref{tab:performance_comparison}) employs a standard accuracy metric from VQA tasks to assess the models' performance in providing correct answers, serving as a direct indicator of their proficiency in processing and answering object counting related questions.

To assess the dataset's VQA annotations, we employed two state-of-the-art VQA models, ViLT \cite{kim2021vilt} and BLIP \cite{li2022blip}, recognized for their robust performance across various VQA benchmarks. These models were queried with counting questions, such as ``How many giraffes are there in the image?'', and their responses were compared against the dataset's annotations. The evaluation, detailed in \cref{tab:performance_comparison}, leverages a specialized accuracy metric tailored for counting tasks within VQA. This metric focuses on the precision of numeric responses to counting queries. In addition to this, we have also reported the mRMSE metric for the VQA task as proposed in Chattopadhyay et al. \cite{Chattopadhyay2017CountEveryday}.

This rigorous approach not only demonstrates OmniCount-191's role in advancing object counting within the VQA framework but also underscores the dataset's capability to challenge and refine the development of VQA models that can navigate the complexities of counting tasks in visual scenes.

In addition to this, we have also demonstrated a qualitative example in Fig~\ref{fig:ground}(a) of how our captioning annotation is useful for object grounding. We can use the captions in the provided OmniCount-191 benchmark to ground fine-grained instances of each of the objects.



\section{Discussions}
\label{sec:discussions}
\subsection{Atomic and Non-atomic Objects}
\begin{figure}[htbp]
\centering
    \includegraphics[width=\linewidth]{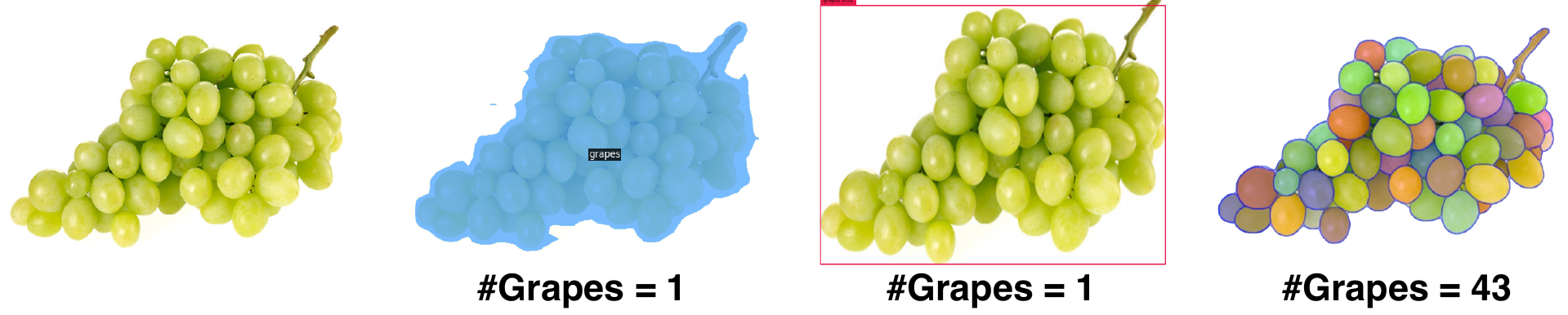}
    \caption{Comparison of model outputs for the query ``grapes'' (left to right): \textbf{instance segmentation} model, \textbf{object detection} model, and our model. While the first two models identify a bunch of grapes as a single entity due to their non-atomic nature, leading to undercounting, our model successfully generates instance-level masks, accurately counting individual grapes.}
    \label{fig:atomic}
\end{figure}

In the main paper, we have mentioned that traditional object detection and instance segmentation models usually fail to enumerate individual instances of a particular class of objects like grapes, berries, bananas, etc. We named those objects as \emph{non-atomic} objects. This term, inspired by the Greek word \emph{atomos} meaning indivisible, applies to objects commonly referenced in aggregate rather than individually. For instance, when prompted with ``grapes'' or ``grape'', Vision-Language Models (VLMs) like CLIP \cite{radford2021learning} tend to identify a bunch of grapes as a single entity, rendering these models ineffective for precise counting tasks (see \cref{fig:atomic}).

To address this limitation, we have developed a strategy utilizing a reference point selection module in conjunction with the Segment Anything Model (SAM). This innovative approach enables the generation of instance-level masks for non-atomic objects, allowing for accurate enumeration of individual instances within a collective entity. By effectively distinguishing between atomic (individually identifiable) and non-atomic (collectively referenced) objects, our method enhances the capability of object counting models to handle a broader range of object types, providing a more nuanced understanding of complex scenes.

\subsection{Is SAM a Good Counter?}
The Segment Anything Model (SAM) \cite{kirillov2023segment} has gained widespread adoption in recent object counting efforts \cite{shi2024training, huang2024point} due to its remarkable zero-shot segmentation capabilities. Trained on an extensive dataset comprising over 1 billion masks and 11 million licensed images, SAM excels at generalizing across a broad spectrum of objects without requiring task-specific training. Its flexibility in accepting various prompts—such as \textit{points}, \textit{boxes}, and \textit{text}—enables it to generate fine-grained segmentation maps, which are particularly useful in object counting tasks.

\begin{figure*}[!h]
\centering
    \includegraphics[width=\linewidth]{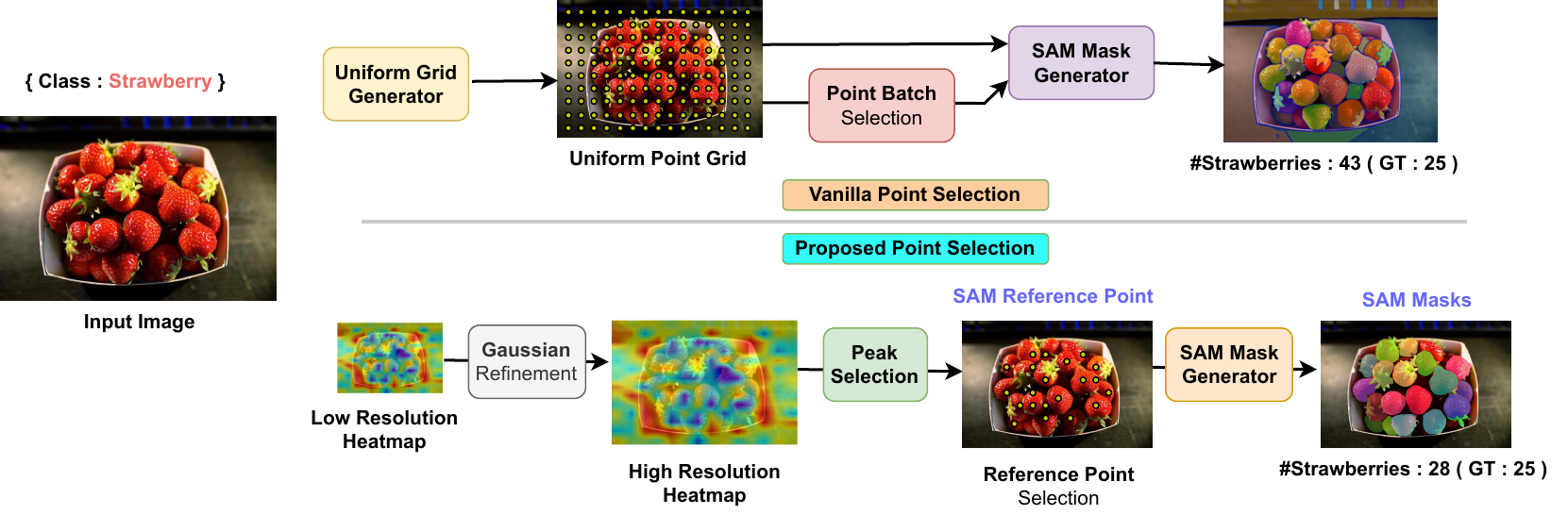}
    \caption{SAM's uniform point placement vs our reference-guided point placement.}
    \label{samvsus}
\end{figure*}

Despite these strengths, SAM has limitations preventing it from matching the latest state-of-the-art counting methods. Key challenges include:

\begin{itemize}
    \item \textbf{SAM is class-agnostic:} SAM is designed to segment objects regardless of their class, which can lead to difficulty distinguishing between different types of objects within a scene. This class-agnostic approach may result in inaccurate counts when specific object categories must be quantified.
    
    \item \textbf{SAM struggles during occlusion:} SAM often struggles to accurately segment objects that are partially obscured or overlapping. This is particularly problematic in dense scenes where objects are closely packed, as it can lead to undercounting or missed detection.
    
    \item \textbf{SAM relies on texture information:} SAM relies heavily on texture and visual features for segmentation. In situations where objects have similar textures, or where there is minimal texture information (such as smooth or featureless surfaces), SAM's performance may decline, resulting in errors in object detection and counting.
    
    \item \textbf{SAM overlays uniform grid:} SAM typically operates by applying a uniform grid across the input image to generate segmentation proposals. While this grid-based approach is efficient, it can struggle with objects of varying sizes or those that do not align well with the grid, potentially resulting in fragmented or incomplete segmentation. Moreover, small and densely packed objects may be missed entirely when fewer grid points are used (e.g., a 32$\times$32 grid). While increasing the number of grid points can mitigate this issue by capturing finer details, it comes at the cost of increased computational resources and the risk of overcounting. This overcounting can occur when points are placed in background areas or on non-target objects, leading to erroneous segmentation results.
\end{itemize}

\noindent \textbf{TFOC} : TFOC \cite{shi2024training} is a training-free object counter that leverages SAM’s segmentation capabilities based on input prompts such as points, boxes, and texts. The model enhances SAM’s segmentation through a novel prior-guided mask generation technique incorporating similarity, segment, and semantic priors. While TFOC effectively detects visually identifiable objects, it struggles in extreme scenarios where objects are too small or heavily occluded, causing them to blend into the background \cite{shi2024training}.

\noindent \textbf{PseCo}: PseCo \cite{huang2024point} introduces a few/zero-shot model that utilizes class-agnostic object localization to generate effective point prompts for SAM. PseCo demonstrates strong performance in base classes but suffers when incorrect exemplars are used for training. Additionally, its reliance on bounding boxes for counting leads to inaccuracies in complex scenes due to occlusion, scale variation, and the inherent limitations of bounding box-based approaches.

\noindent \textbf{OmniCount}: OmniCount distinguishes itself from previous SAM-based object counters by offering a truly open-vocabulary, training-free approach that leverages semantic and geometric cues from pre-trained models. Unlike SAM, which applies a uniform grid for point placement, OmniCount introduces a reference-guided point placement strategy that significantly reduces overcounting and undercounting. This approach is particularly effective in handling dense scenes, where SAM’s uniform grid may falter. OmniCount’s method ensures that reference points are optimally placed, improving segmentation accuracy and making it more reliable for complex object counting tasks, including those involving varying object sizes, as shown in \cref{samvsus}. In essence, while SAM-based models like TFOC and PseCo have paved the way for leveraging SAM in object counting, OmniCount represents a paradigm shift by addressing the limitations of these earlier models. OmniCount offers a more robust solution for accurate object counting across diverse and complex scenarios by avoiding the pitfalls of uniform grid placement and bounding box dependency.

\subsection{Referring Expression vs Multi-label Counting}
We assess the performance of OmniCount in the Referring Expression Counting (REC) task using the REC-8K benchmark, as presented in \cref{tab:recomni}. In this evaluation, we utilize referring expression text as input prompts for OmniCount, replacing the standard class text. The competing methods include ZSOC \cite{xu2023zero}, TFOC \cite{shi2024training}, CounTX \cite{amini2023open}, GroundingDino \cite{liu2023grounding}, and GrREC \cite{Dai2024REC}. As shown in \cref{tab:recomni}, OmniCount outperforms most existing methods on the REC task. It is also the best-performing model among other training-free alternatives. Although GrREC exhibits slightly better performance, it is important to note that GrREC is specifically optimized for the REC task. OmniCount, on the other hand, is designed to be a more versatile model, handling a broader range of counting tasks. This versatility may introduce minor trade-offs in tasks like REC, where task-specific optimization, as seen in GrREC, can lead to marginally higher accuracy. Nevertheless, OmniCount’s strong results in REC, despite its broader focus, underscore its robustness and adaptability across diverse scenarios. However, our model can accommodate referring expression as input if we replace the semantic estimation module with a more generic referring semantic segmentation module. Thus it can be an interesting future research direction.  

\begin{table}
\centering
\caption{Performance of various Object Counting and Referring Expression Counting (REC) models on the REC-8K benchmark.}
\label{tab:recomni}
\resizebox{\linewidth}{!}{%
\begin{tabular}{lcccccc} 
\hline
\textbf{Method} & \textbf{Training} & \textbf{MAE$\downarrow$} & \textbf{RMSE$\downarrow$} & \textbf{Prec$\uparrow$} & \textbf{Rec$\uparrow$} & \multicolumn{1}{l}{\textbf{F1$\uparrow$}} \\ 
\hline
 ZSOC \cite{xu2023zero} & \cmark & 14.93 & 29.72 & - & - & - \\
 TFOC \cite{huang2024point} & \xmark & 12.77 & 32.68 & 0.23 & 0.07 & 0.11 \\
 CounTX \cite{amini2023open} & \cmark & 11.84 & 25.62 & - & - & - \\
 GroundingDino \cite{liu2023grounding} & \xmark & 11.71 & 26.97 & 0.59 & 0.25 & 0.35 \\
 GrREC \cite{Dai2024REC} & \cmark & \underline{6.50} & \underline{19.79} & \underline{0.67} & \underline{0.72} & \underline{0.69} \\
 \rowcolor[rgb]{0.753,0.753,0.753} \textbf{OmniCount} & \xmark & \textbf{7.44} & \textbf{20.88} & \textbf{0.61} & \textbf{0.54} & \textbf{0.57} \\
\hline
\end{tabular}
}
\end{table}



\section{Limitations and Future Work}
\label{weakness}
\begin{figure*}[!h]
\centering
    \includegraphics[width=0.9\linewidth]{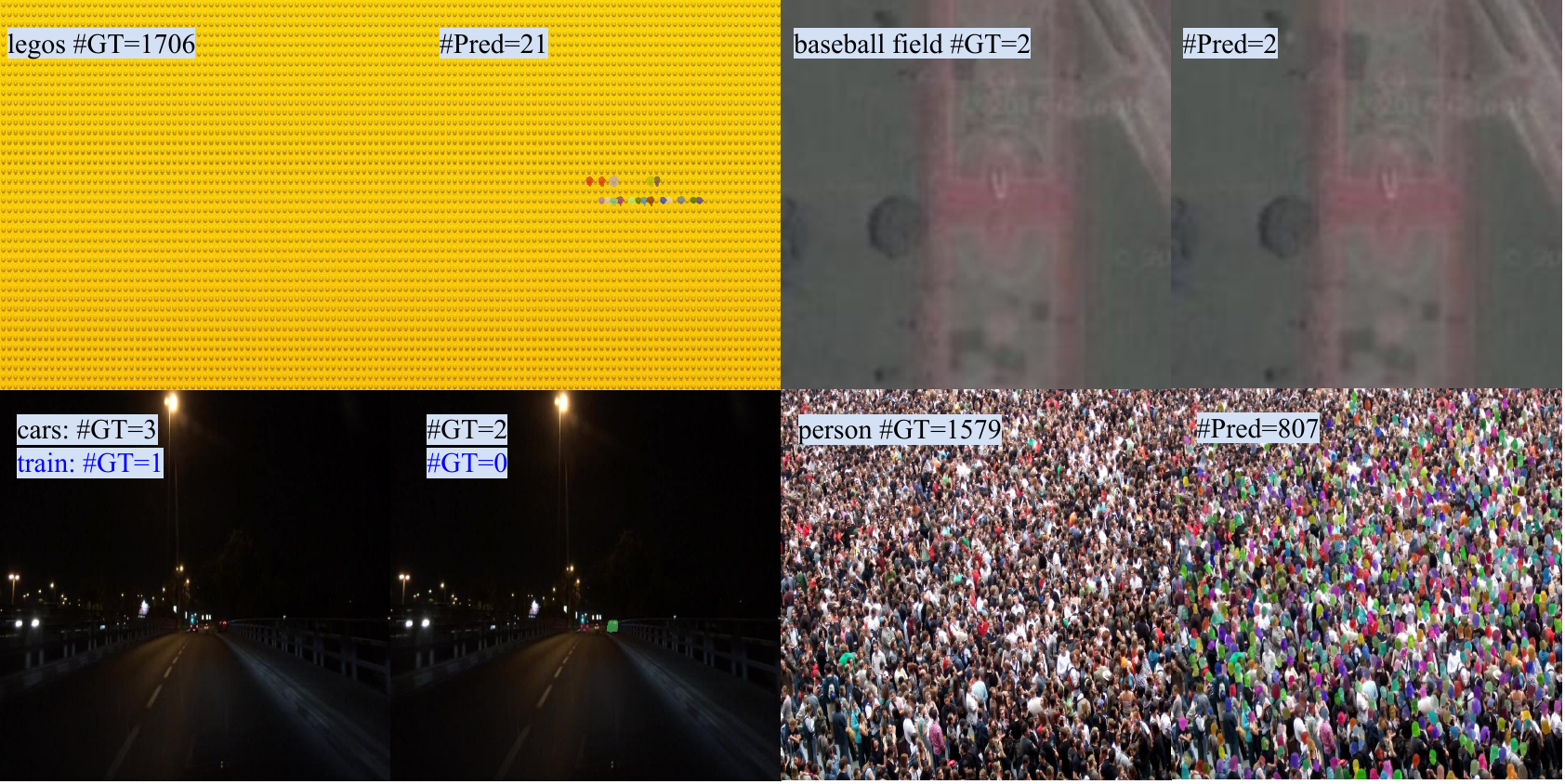}
    \caption{OmniCount struggles to count objects in images having: low-resolution (legos), large distance from camera (baseball field), low illumination (cars/train), and crowd scenarios (person).}
    \label{fig:fail}

\end{figure*}

In this section, we identify and discuss three primary limitations encountered by our OmniCount, alongside their implications for its performance:

\myparagraph{Low-Resolution and Low Illumination Images}: OmniCount encounters challenges in accurately generating object proposals in low-resolution images, primarily due to the inherent limitations of current semantic segmentation models, which struggle to capture fine details necessary for precise object identification. As a result, the quality of the input images plays a crucial role in OmniCount’s object-counting accuracy. Additionally, the model’s performance is affected under very low illumination, where insufficient lighting impairs object detection and counting processes. These observations underscore the importance of ensuring adequate image resolution and lighting conditions to maximize OmniCount’s effectiveness. To mitigate these issues, integrating image enhancement techniques or pre-processing steps could improve the model’s performance in challenging conditions.



\myparagraph{Distance from Objects}: Performance issues arise when the camera is positioned far from the target objects. In these instances, the depth estimation model, a critical component of OmniCount, fails to gauge objects' distance and dimensions accurately. Consequently, the overall effectiveness of the model is limited by the performance of the semantic and point selection components in such conditions. However, these limitations can be mitigated by incorporating dynamic zoom-in networks \cite{gao2018dynamic}, which enhance the model’s ability to focus on distant objects and improve accuracy.

\myparagraph{Crowd counting failure}: {While OmniCount excels at multi-label counting across diverse categories, it faces challenges in crowd counting, particularly in densely populated areas. Similar to previous works \cite{huang2024point, shi2024training}, crowd counting poses unique challenges that often require specialized techniques like density estimation,
where each pixel can have one object instance. Such detailed localization of objects is only possible with training-based methods \cite{wan2024robust,pelhan2024dave} which is unlike our proposed training-free setting. OmniCount uses object features as a hint to detect the possible location of objects that may overlook very densely packed or overlapping objects, and thus may not work well for extremely dense scenarios like crowd counting where significant scale variations and the lack of specific crowd-focused training reduce its effectiveness. In general, the existing training-free object counting methods uniformly share the same drawbacks of inferior counting ability in dense scenarios. However, it’s important to note that OmniCount performs effectively in moderately crowded scenes (figure 7 of the main paper), particularly when the number of individuals is below a certain threshold ( $>$ 60\% accuracy for $<$ 100 instances, shown in figure 9 of main paper). In these scenarios, OmniCount can accurately estimate counts without requiring additional methods. This is possible as we separate occluded objects at different depths to extract the object features thus allowing it to be counted. As a result, our method performs better than existing training-free counting methods in dense scenarios. Incorporating specialized crowd-counting techniques could enhance performance to handle these more challenging situations. We illustrate all the difficult cases in \cref{fig:fail}}.

\section{Additional Visualizations}
\label{fig:vizs}

We provide domain-level visualizations from Omnicount on our Omnicount-191 dataset.

\begin{figure*}[!ht]
\centering
\includegraphics[width=\linewidth]{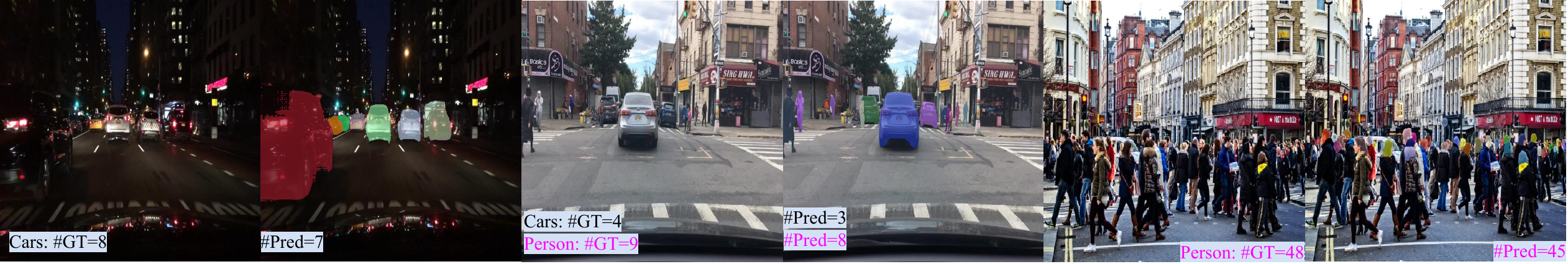}
\caption{\textbf{Urban}}
\vspace{3pt}
\label{fig:urban}
\end{figure*}

\begin{figure*}[!ht]
\centering
\includegraphics[width=\linewidth]{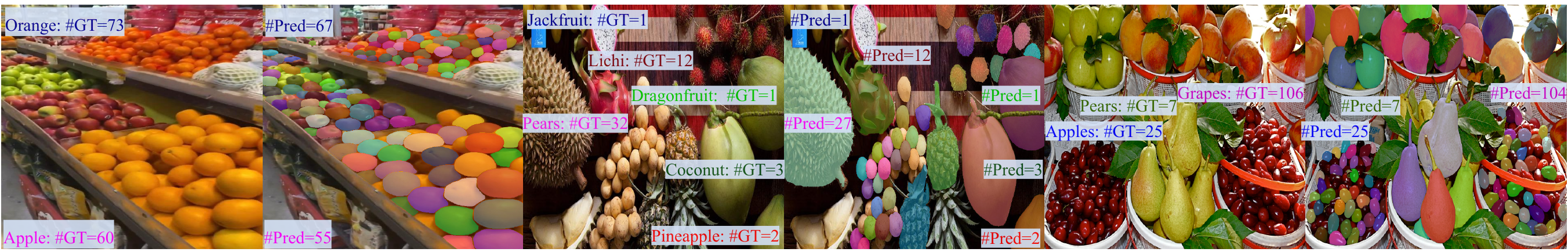}
\caption{\textbf{Fruits}}
\label{fig:fruits}
\end{figure*}

\begin{figure*}[!ht]
\centering
\includegraphics[width=\linewidth]{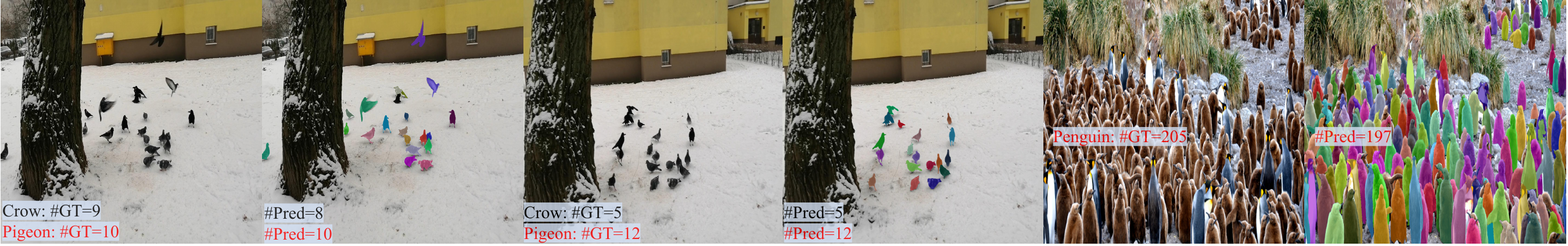}
\caption{\textbf{Birds}}
\vspace{5pt}
\label{fig:birds}
\end{figure*}

\begin{figure*}[!ht]
\centering
\includegraphics[width=\linewidth]{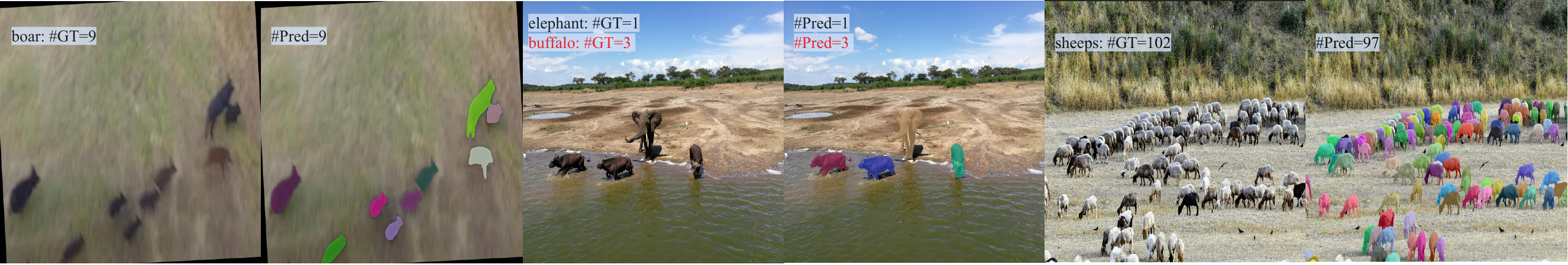}
\caption{\textbf{Wild}}
\label{fig:wild}
\end{figure*}

\begin{figure*}[!ht]
\centering
\includegraphics[width=\linewidth]{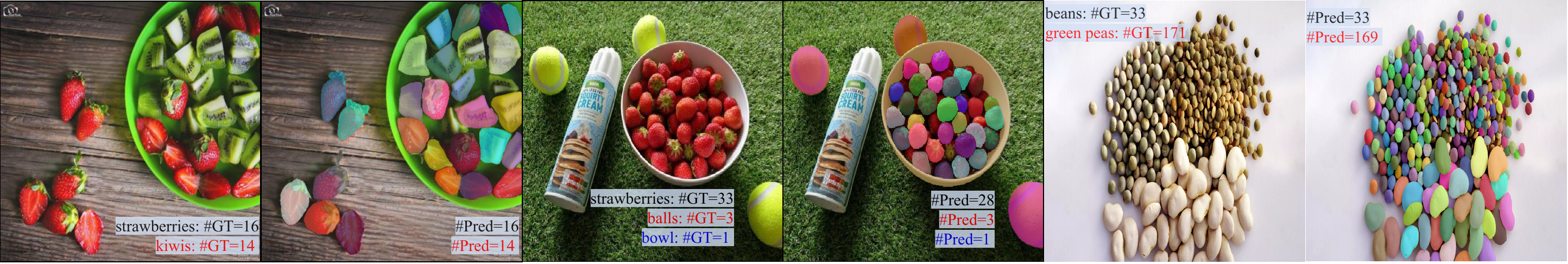}
\caption{\textbf{Household}}
\label{fig:house}
\end{figure*}

\begin{figure*}[!h]
\centering
\includegraphics[width=\linewidth]{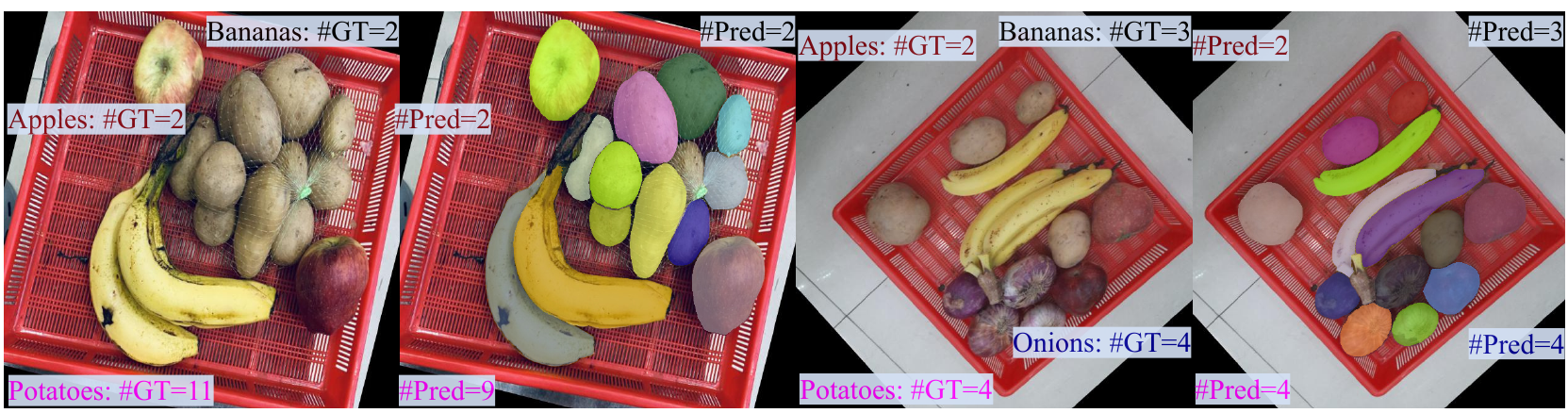}
\caption{\textbf{Agriculture}}
\label{fig:agri}
\end{figure*}

\begin{figure*}[!ht]
\centering
\includegraphics[width=\linewidth]{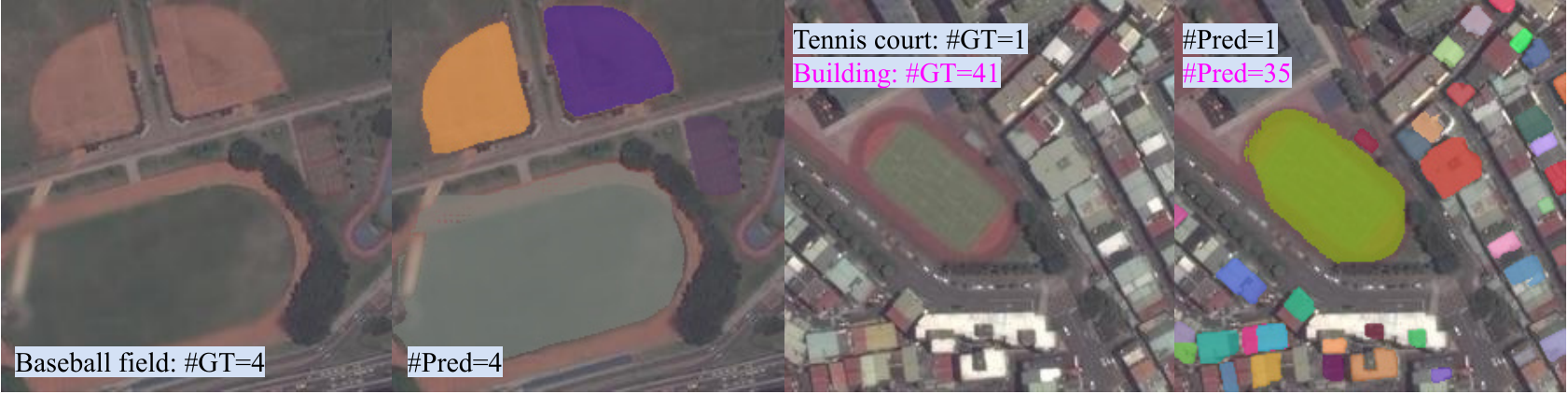}
\caption{\textbf{Satellite}}
\label{fig:satellite}
\end{figure*}

\newpage

\begin{figure*}[!h]
\centering
\includegraphics[width=\linewidth]{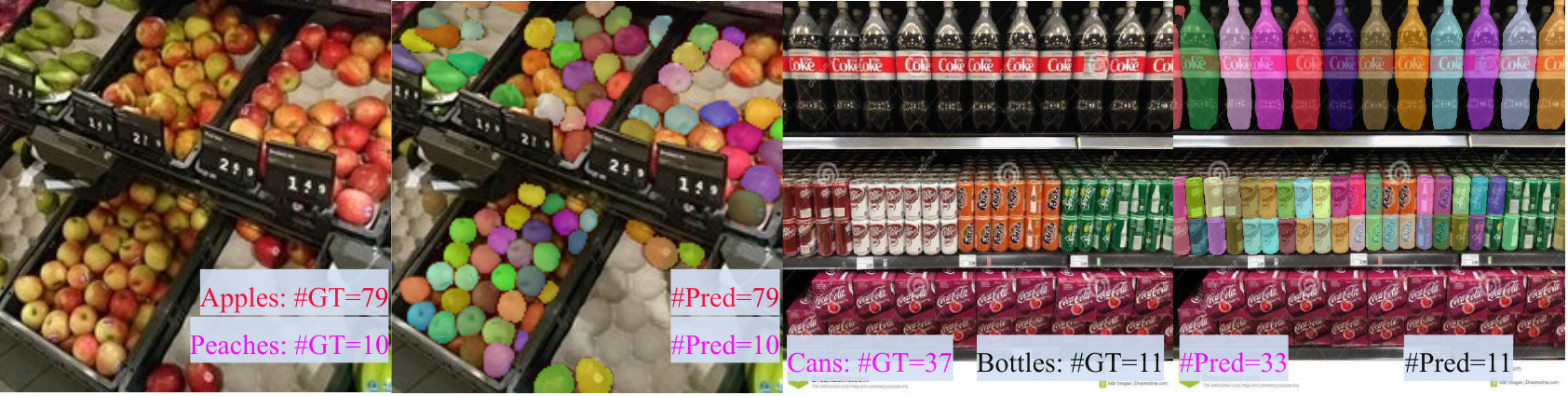}
\caption{\textbf{Supermarket}}
\label{fig:supermarket}
\end{figure*}

\vspace{-3mm}

\begin{figure*}[!h]
\centering
\includegraphics[width=\linewidth]{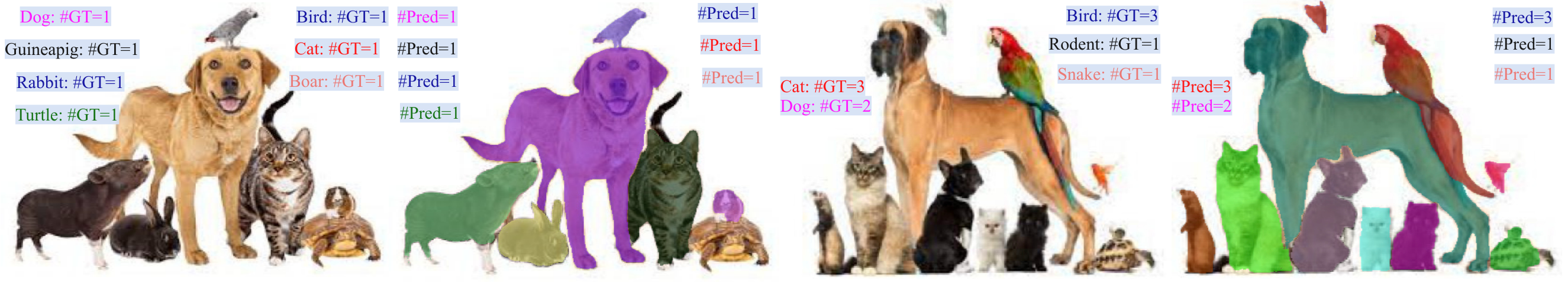}
\caption{\textbf{Pets}}
\label{fig:pets}
\end{figure*}

\newpage

\bibliography{aaai25}
\end{document}